\def\year{2021}\relax
\documentclass[letterpaper]{article} 
\usepackage{aaai21}  
\usepackage{times}  
\usepackage{helvet} 
\usepackage{courier}  
\usepackage[hyphens]{url}  
\usepackage{graphicx} 
\usepackage{natbib}  
\usepackage{caption} 
\usepackage{amsthm}
\usepackage{amsmath,amssymb} 
\usepackage{booktabs}       
\usepackage{rotating}
\usepackage{multirow}
\newcommand{\norm}[1]{\left\lVert#1\right\rVert}
\newtheorem{assumption}{Assumption}
\newtheorem{theorem}{Theorem}
\newtheorem{definition}{Definition}
\usepackage{subcaption}
\usepackage{bm}
\usepackage{enumitem}
\usepackage{thmtools}
\usepackage{thm-restate}

\let\given\givenbase
\usepackage{mathtools}
\usepackage{microtype}      

\title{Noise Estimation Using Density Estimation \\for Self-Supervised Multimodal Learning}
\author{
    Elad Amrani\textsuperscript{\rm 1,2},
    Rami Ben-Ari\textsuperscript{\rm 1},
    Daniel Rotman\textsuperscript{\rm 1},
    Alex Bronstein\textsuperscript{\rm 2}
}
\affiliations{

    \textsuperscript{\rm 1}IBM Research AI,
    \textsuperscript{\rm 2}Technion

}

\begin{document}

\maketitle

\begin{abstract}
One of the key factors of enabling machine learning models to comprehend and solve real-world tasks is to leverage multimodal data. Unfortunately, annotation of multimodal data is challenging and expensive. Recently, self-supervised multimodal methods that combine vision and language were proposed to learn multimodal representations without annotation. However, these methods often choose to ignore the presence of high levels of noise and thus yield sub-optimal results. In this work, we show that the problem of noise estimation for multimodal data can be reduced to a multimodal density estimation task. Using multimodal density estimation, we propose a noise estimation building block for multimodal representation learning that is based strictly on the inherent correlation between different modalities. We demonstrate how our noise estimation can be broadly integrated and achieves comparable results to state-of-the-art performance on five different benchmark datasets for two challenging multimodal tasks: Video Question Answering and Text-To-Video Retrieval. Furthermore, we provide a theoretical probabilistic error bound substantiating our empirical results and analyze failure cases. Code: \url{https://github.com/elad-amrani/ssml}.
\end{abstract}

\section{Introduction}
Multimodal learning is a well established methodology for tackling complex and challenging artificial intelligence tasks such as Visual Question Answering \cite{antol2015vqa,jang2017tgif,gao2018motion,xu2017video,fan2019heterogeneous} and Text-to-Video Retrieval \cite{Liu19a,mithun2018learning,yu2018joint,miech2018learning,song2019polysemous}. The motivation for gleaning information from multiple correlated data sources comes from how we as humans perceive the world and learn from experience. Using the correlation between speech and vision, a person is able to recognize objects by their names while learning the visual characteristics. Additionally, concepts can be learned separately and a combination can be comprehended automatically, e.g., `running' and `beach' vs. `running on the beach'.

Manual annotation of large-scale datasets and specifically multimodal datasets is challenging and expensive. This difficulty results in a shortage which limits the progress of supervised machine learning and has become the key development bottleneck. Recently, to combat costs and effort of annotation, self-supervised machine learning \cite{zhang2016colorful,noroozi2016unsupervised,pathak2016context,misra2016shuffle,wei2018learning,vondrick2016generating,srivastava2015unsupervised} presents new ways to better utilize the abundant unlabeled data on the web. However, most self-supervised systems aim to learn from a single data modality, which limits their applicability. 

In contrast to the above, \cite{miech2019howto100m,miech2020end,amrani2019learning,moriya2019grounding,sun2019videobert,sun2019contrastive} recently showed that unlabeled instructional videos could be used as training data for a self-supervised multimodal learning system due to the high correlation between the spoken word and the ongoing visuals. Unfortunately, such systems are forced to deal with high noise levels and thus yield sub-optimal results as we show in this paper.

In this paper, we propose a novel noise robust multimodal representation learning building block for self-supervised learning. We utilize the inherent correlation between different modalities for efficient multimodal learning in the presence of extreme levels of noise. Specifically, we show that noise estimation can be reduced to a density estimation problem. We define a multimodal similarity function and show that based on this function, noise is correlated with sparsity and vice versa.

Ultimately, we integrate our proposed building block into an embedding model and learn superior joint video-text representations that achieve comparable state-of-the-art performance on five datasets: MSRVTT \cite{xu2016msr}, LSMDC \cite{rohrbach2015dataset}, MSVD \cite{chen2011collecting}, MSRVTT-QA \cite{xu2017video} and MSVD-QA \cite{xu2017video}; for two different tasks: Video Question Answering and Text to Video Retrieval. Additionally, we provide a theoretical probabilistic error bound substantiating our empirical results and analyze failure cases.
    
\textbf{Contributions.} The key contributions of this paper are four fold: 
\begin{enumerate}[noitemsep,topsep=0pt, leftmargin=*]
    \item We show that the problem of noise estimation for multimodal data can be efficiently reduced to a multimodal density estimation task. 
    \item We propose a novel building block for noise-robust multimodal representation learning and demonstrate its integration into the max margin ranking loss function.
    \item We demonstrate comparable state-of-the-art performance on five datasets for two different challenging multimodal tasks by utilizing our approach for self-supervised multimodal learning with the HowTo100M dataset \cite{miech2019howto100m}.
    \item We substantiate our empirical results with a theoretical analysis of the proposed method that includes a probabilistic error bound.
\end{enumerate}

\section{Related Work}

\paragraph{Self-Supervised Learning.}
Self-supervised learning methods strive to learn informative data representations by defining and solving a pretext task. In these tasks, pseudo labels can be generated automatically and compact data representations must be learned in order to solve these tasks. Many pretext tasks were proposed in recent years: colorizing grayscale images \cite{zhang2016colorful}, image jigsaw puzzle \cite{noroozi2016unsupervised}, image inpainting \cite{pathak2016context}, video frame order verification \cite{misra2016shuffle}, video future prediction \cite{vondrick2016generating,srivastava2015unsupervised}, audio-visual correspondence \cite{korbar2018cooperative,arandjelovic2017look}, speech-visual correspondence \cite{miech2019howto100m,miech2020end,amrani2019learning,moriya2019grounding,sun2019videobert,sun2019contrastive}, etc.
In this work, we focus on speech-visual correspondence in unlabeled instructional videos, where speech is converted to text using an automatic speech recognition system. Speech-visual correspondence is considered a difficult pretext task due to extremely noisy pseudo labels, yet it can be a highly advantageous task since it provides semantic information of visual features in the form of natural text. Such valuable information can be utilized to solve many challenging multimodal downstream tasks as we show in Section \ref{experiments}.

\paragraph{Multimodal Representation Learning.}
The word {\it modality} refers to a particular form of sensory perception, such as the visual and auditory modalities. A machine learning task or dataset is said to be {\it multimodal} when it includes a number of modalities. 
Multimodal representation learning frameworks can be divided into three types: (a) joint representation which aims to learn a shared semantic subspace \cite{salakhutdinov2009deep,srivastava2012learning,antol2015vqa}; (b) an encoder-decoder framework which aims to translate from one modality into another and keep their semantics consistent \cite{mao2014deep,rohrbach2015long,venugopalan2015translating,reed2016generative}; and (c) coordinated representation which aims to learn separated yet coordinated representations for each modality under some constraints \cite{weston2010large,frome2013devise,socher2014grounded,wang2016learning,ivan2016order,tian2019contrastive}.
In this work, we focus on coordinated representations that enforce similarity among them. Our goal is to enforce the multimodal representations of similar `concepts' to be close to each other. E.g., a video of a man running on the beach should be close in representation to the textual representation of `a man running on the beach' as opposed to `a man cooking in the kitchen'.
The multimodal representation described above is highly valuable for solving multimodal machine learning tasks. If a machine learning model learns to link between the visuals and text of specific concepts it should be able, for example, to answer natural language questions about visual content, or do cross-modal retrieval more easily (Section \ref{ds_tasks}).

\paragraph{Density Estimation.}
The aim of density estimation is to estimate the probability density function underlying the data, which is assumed to be $i.i.d$. Existing density estimation algorithms can be divided into two categories: (a) parametric or semi-parametric approaches such as Gaussian Mixture models \cite{mclachlan2007algorithm,wang2015nonparametric} and probabilistic graphical models \cite{koller2009probabilistic}; and (b) non-parametric approaches such as histograms, Splines \cite{stone1994use}, neural network-based density estimation \cite{uria2014deep,papamakarios2017masked} and Kernel Density Estimation \cite{silverman2018density,terrell1992variable}. For an extended review on density estimation for high-dimensional data see \cite{wang2019nonparametric}.
In this work, we utilize {\it multimodal} k-Nearest Neighbor density estimation, which is a special case of Kernel Density Estimation. With it, we form a novel noise-robust multimodal representation learning model.

\paragraph{Learning with Noisy Data.}
Learning with noisy data can be divided into two approaches: (a) formulating explicit or implicit noise models to characterize the distribution of noisy and true labels using neural networks \cite{Goldberger2017TrainingDN,lu2018mentornet,sukhbaatar2015training}, graphical models \cite{xiao2015learning,li2017learning}, etc.  and (b) using correction methods. E.g., relabeling the data during training \cite{reed2015training}, jointly optimizing the model's parameters and estimating true labels \cite{tanaka2018joint}, using noise-tolerant loss function \cite{ghosh2017robust,van2015learning} or noise tolerant training algorithms \cite{li2019learning}. However, these methods do not deal with multimodal association label (Definition \ref{association_def}) and often require a small set of data with clean labels to be available.
In this work, we propose a true label estimation method for multimodal data that does not require availability of clean labels. We base our estimation on the correlation between modalities alone. 

\section{Method}
\subsection{Motivation}
In multimodal data, a sample is said to be noisy when two or more modalities do not share the same semantic meaning. For example, a video-text pair that is associated with each other, yet the text is not related to the ongoing visuals. Existing multimodal embedding models are susceptible to such noisy data, i.e., the model is likely to adjust itself to the noise in the data and thus yield sub-optimal results. This scenario is very common in the case of self-supervised multimodal learning and even when learning from unlabeled instructional videos. Although in these instructional videos there is some correlation between caption (speech transcription) and vision, unfortunately often a person is talking about something that is not present visually. For example, in the HowTo100M dataset \cite{miech2019howto100m}, the authors manually inspected $400$ randomly sampled clip-caption pairs and found that in about half there was not a single object or action mentioned in the caption that was also visually present in the video clip.
\begin{figure*}[tp]
\begin{subfigure}[b]{0.65\textwidth}
    \centering
    \includegraphics[width=1\linewidth]{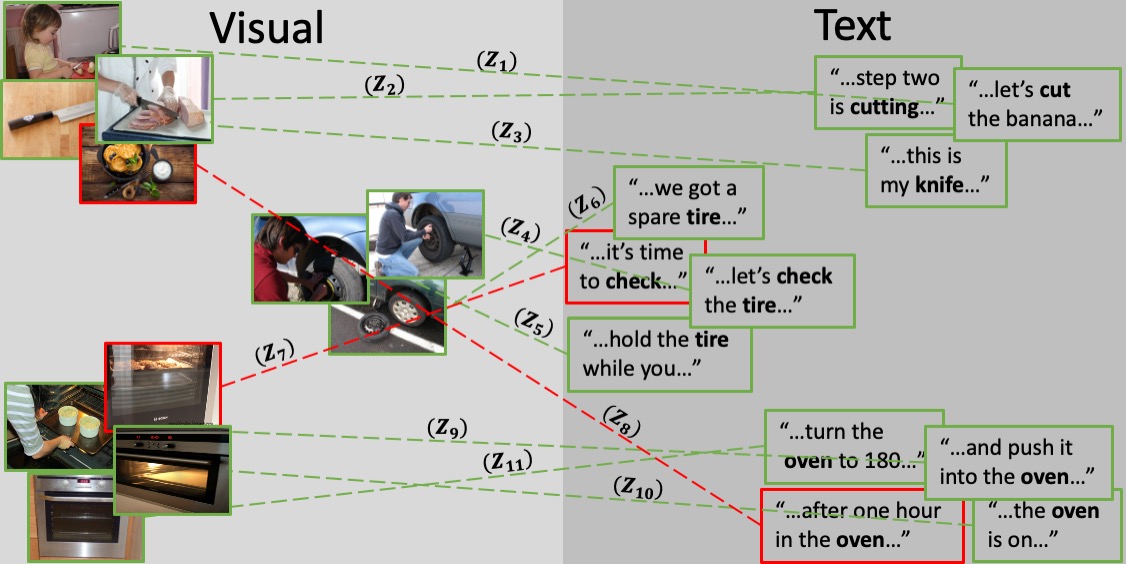}
    \caption{\textbf{Multimodal data visualization}. Each initial monomodal embedding space contains somewhat dense clusters of `concepts', where a `concept' could be a specific object or action (e.g., `cutting', `knife', `check', `tire', `oven', etc.). It is likely that {\it correctly associated} (Definition \ref{association_def}) pairs form dense clusters in both modalities that contain pairs that are also associated with each other and of the same `concept' (GREEN, $z_1$ - $z_6$, $z_9$ - $z_{11}$). In contrast, a {\it wrongly associated} (Definition \ref{association_def}) pair may still belong to dense clusters in both modalities but those clusters are not likely to contain pairs that are associated with each other (RED, $z_7$ and $z_8$). Best viewed in color.}
    \label{fig:visualization_a} 
\end{subfigure}
\hfill
\begin{subfigure}[b]{0.33\textwidth}
    \centering
    \includegraphics[width=0.985\linewidth]{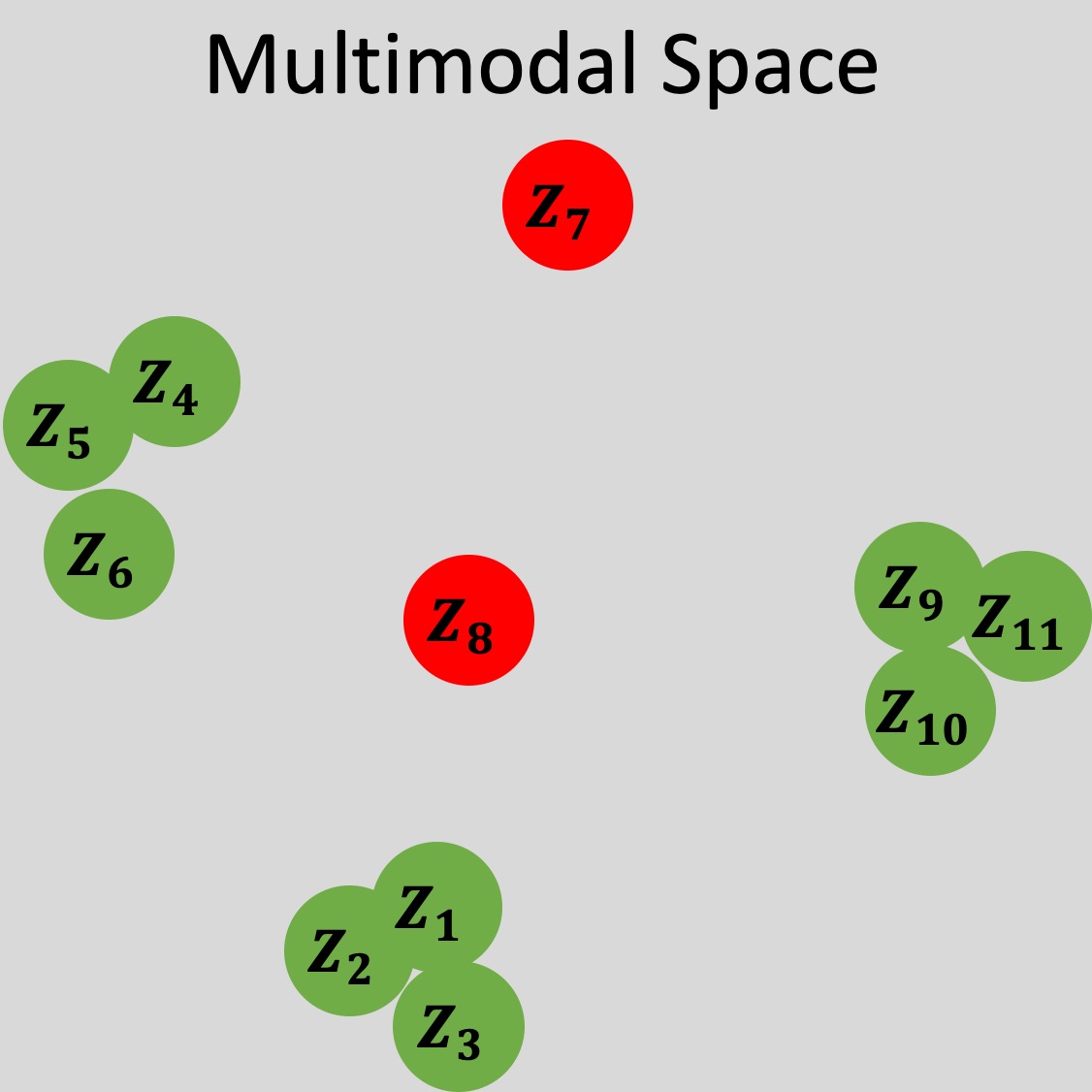}
    \caption{\textbf{Multimodal space defined by \eqref{eq:dist}}. Each point above ($\{z_i\}_{i=1}^{11}$) represents a single pair from the left sub-figure. The distance between points that is visualized is computed based on \eqref{eq:dist}. Given Assumption \ref{assume}, in the multimodal space above, {\it correctly associated} pairs are correlated with high density and vice-versa. Best viewed in color.}
    \label{fig:visualization_b}
\end{subfigure}
\caption{Noise estimation using multimodal density estimation}
\label{fig:visualization}
\end{figure*}
To deal with noise, we suggest to utilize the inherent correlation between the different modalities that is based on the Definition and Assumption below. See Fig. \ref{fig:visualization} for a visualization and a detailed explanation. 

\begin{definition}
    \label{association_def}
    A correctly (wrongly) associated pair is a clip-caption pair $(v, c)$ that share (do not share) the same semantic meaning or concept, i.e., the caption $c$ describes (does not describe) the ongoing visuals $v$.
\end{definition}

\begin{assumption}[Mixture Model]
    \label{assume}
    The distributions of the videos and captions can be represented using a general mixture model of $T$ components in the corresponding modality. Denoting by $a,b \in \{1,\dots,T\}$, respectively, the concept to which the video $v$ and the caption $c$ belong, we can write $v|a \sim p_{\mathrm{v}}(v|a)$ and $c|b \sim p_{\mathrm{c}}(c|b)$.
\end{assumption}

If Assumption \ref{assume} holds, then {\it correctly associated} pairs form dense clusters in both modalities that contain pairs that are also associated with each other (see Fig. \ref{fig:visualization_a}). Thus, by defining a multimodal similarity function (i.e., a similarity measure between pairs), we can formulate the task of finding {\it correctly associated} pairs simply as a multimodal density estimation task. In this formulation, pairs in dense areas will be more likely to be {\it correctly associated}, while pairs in sparse areas will be more likely to be {\it wrongly associated} (see Fig. \ref{fig:visualization_b}).

\subsection{Notation and Problem Formulation}
Let $\{ (v_i,c_i) \in \mathbb{R}^{d_{\mathrm{v}}} \times \mathbb{R}^{d_{\mathrm{c}}}\}_{i=1}^M$ denote the set of clip-caption pairs, where for each $i$, the video clip $v_i$ is associated with the caption sentence $c_i$, and $M$ denotes the size of the dataset. Let $p_i \in \{0, 1\}$ denote a binary indicator for whether the pair $(v_i, c_i)$ is {\it correctly associated} ($p_i = 1$) or {\it wrongly associated} ($p_i = 0$).
Let $f_{\mathrm{v}} : \mathbb{R}^{d_{\mathrm{v}}} \rightarrow \mathbb{R}^d$ and $f_{\mathrm{c}} : \mathbb{R}^{d_{\mathrm{c}}} \rightarrow \mathbb{R}^d$ denote the embedding functions of the videos and the captions, respectively, into a common representation space.
The task of {\it noise robust} multimodal representation learning aims to map all of the data modalities to a single embedding space such that for all $v_i$ that is {\it correctly associated} with $c_i$, $f_{\mathrm{v}}(v_i) \approx f_{\mathrm{c}}(c_i)$ in the sense of some similarity function.
\subsection{Noise Estimation Using Multimodal Density Estimation}
\label{mmde}
For the ease of notation, we will denote the pair as $z_i = (v_i, c_i)$. Let us define a similarity function between pairs, $S : \mathbb{R}^{d_{\mathrm{v}}+d_{\mathrm{c}}} \times \mathbb{R}^{d_{\mathrm{v}}+d_{\mathrm{c}}} \rightarrow \mathbb{R}$.
\begin{equation}
    \label{eq:dist}
    S(z_i, z_j) \triangleq \min\bigg\{\frac{s(v_i, v_j) -  \bar{\mu}_{\mathrm{v}} }{\bar{\sigma}_{\mathrm{v}}}, \frac{s(c_i, c_j) -  \bar{\mu}_{\mathrm{c}} }{\bar{\sigma}_{\mathrm{c}}} \bigg\},
\end{equation}
where $s$ can be, for example, the cosine similarity function $s(x, y) = \frac{x^{\intercal}y}{\norm{x}\norm{y}}$; $\bar{\mu}_{\mathrm{v}}, \bar{\mu}_{\mathrm{c}}$ and $\bar{\sigma}_{\mathrm{v}}, \bar{\sigma}_{\mathrm{c}}$ are the sample means and standard deviations of each modality, i.e., the similarity values of each modality are normalized before taking the minimum.  Using \eqref{eq:dist}, a pair $z_i$ is close to $z_j$ only if $v_i$ is close to $v_j$ and $c_i$ is close to $c_j$ as well. 

We denote by $\hat{p}_i$ the estimated probability of $z_i$ being {\it correctly associated}, and compute it using its local $k$-NN density estimation normalized such that $\hat{p}_i \in [0, 1]$:
\begin{equation}
    \label{eq:p_i}
    \hat{p}_i \triangleq \frac{\Bar{S}_i - \min(\{\Bar{S}_i\}_{i=1}^M)}{\max(\{\Bar{S}_i\}_{i=1}^M) - \min(\{\Bar{S}_i\}_{i=1}^M)},
\end{equation}
where,
\begin{equation}
    \label{eq:S_i}
    \Bar{S}_i = \frac{1}{K}\sum_{k=1}^K{S(z_i, z_{i_k})}, i \in [M],
\end{equation}
$z_{i_k}$ is the $k$-th nearest neighbor of $z_i$ and $S$ is the similarity function defined in \eqref{eq:dist}.

\subsection{Soft Max Margin Ranking Loss}
\label{subsection:soft_max_margin}
Integrating our noise estimation component from above into a max margin ranking loss function \cite{wang2014learning,schroff2015facenet} is straightforward. We weight each pair $z_i$ with its fixed estimated probability $\hat{p}_i$ of being {\it correctly associated}. We call it {\it Soft} Max Margin Ranking:
\begin{multline}
\label{eq:soft_max_margin}
    L_{soft-rank} = \sum_{i \in P}\bigg({\hat{p}_i}\sum_{j \in N_i}{\max\{0, s_{ij} - s_{ii} + \delta\}} + \\\max\{0, s_{ji} - s_{ii} + \delta\}\bigg),
\end{multline}
where, $P$ is the set of noisy associated (positive) pairs, $N_i$ is the set of negative pairs for clip-caption pair $(v_i, c_i)$, $\hat{p}_i$ is defined in \eqref{eq:p_i}, $s_{ij}$ is the similarity score between the embedding of the clip-caption pair $(f_{\mathrm{v}}(v_i), f_{\mathrm{c}}(c_j))$, and $\delta$ is the margin. The first term in the equation above is for matching a video with a negative caption and the second term is for matching a caption with a negative video.

We note that a different integration approach is to discard samples with $\hat{p}_i$ below a certain threshold. However, for real-data experiments we found the performance to be substantially worse. 
\section{Theoretical Analysis}
\label{theoretical_analysis}
We present a theoretical probabilistic error upper bound of our noise estimation approach. For simplicity we assume the data is distributed under a Gaussian Mixture model.
\subsection{Probabilistic Error Upper Bound}
\begin{theorem}
\label{theorem:error_bound}
Let $Z=(X,Y) \in \mathbb{R} \times \mathbb{R}$ be a random pair of scalars satisfying Assumption \ref{assume} for a Gaussian mixture of $T > 1$ equi-probable concepts. Denoting by $(A,B) \in \{1,\dots,T\}^2$ the concepts to which the pair $Z$ belongs, $X|A=a \sim \mathcal{N}(\mu_a,\sigma_a^2)$ and $Y|B=b \sim \mathcal{N}(\mu'_b,\sigma_b^{'2})$. We further assume that each component of the mixture is $6\sigma$-separated, i.e., $|\mu_i - \mu_j| > 6\sigma_{\mathrm{max}}$ and $|\mu'_i - \mu'_j| > 6\sigma'_{\mathrm{max}}$ for every $i \neq j$, where $\sigma_{\mathrm{max}} \triangleq \max_{t}\{\sigma_t\}$ and $\sigma'_{\mathrm{max}} \triangleq \max_{t}\{\sigma'_t\}$.

Let $Pr(A=a \ne B=b) = \frac{\eta}{T(T-1)}$ and $Pr(A=B=a) = \frac{1-\eta}{T}$ for every $a,b \in \{1,\dots,T\}$. The binary indicator $P$ denoting whether the pair $(X,Y)$ is correctly associated consequently has $Pr(P=1) = Pr(A=B) = 1-\eta$, where $\eta$ is the noise ratio of the dataset.

Let $\{ z_i = (x_i,y_i) \}_{i=1}^M$ be a finite sample of pairs drawn independently from the described model, and let $\Bar{S}_i$ be
the average similarity between $z_i$ and its $K$ nearest neighbors as defined in \eqref{eq:S_i}, with $S(z_i, z_j) \triangleq \frac{s(x_i, x_j) + s(y_i, y_j)}{2}$, and $s(x, x') \triangleq -|x - x'|$.
Then, the following bounds hold for every $\tau$ and $t>0$,
\begin{equation}
    \label{eq:S_i_bar_proof1}
    P(\Bar{S_i} \geq \tau \given p_i = 0) \leq \frac{\mathcal{M}_{\Bar{S}_i \given p_i = 0}(t)}{e^{t \tau}},
\end{equation}
\begin{equation}
    \label{eq:S_i_bar_proof2}
    P(\Bar{S_i} \leq \tau \given p_i = 1) \leq \frac{\mathcal{M}_{\Bar{S}_i \given p_i = 1}(-t)}{e^{-t \tau}},
\end{equation}
where, $\mathcal{M}_{\Bar{S}_i \given p_i}(t)$ is the moment generating function of $\Bar{S}_i \given p_i$ defined in Appendix \ref{proof_of_theorem}, Eq. \eqref{eq:mgf_S_i_bar}.
\end{theorem}
The proof is provided in Appendix \ref{proof_of_theorem}.
It is important to remark that for the simplicity of analysis we assumed the pairs to be formed of scalars. While, from the first glance, this assumption might severely limit the possible configurations of the concepts in each of the modalities, in our analysis we made no assumptions whatsoever on the way the concepts are collocated in space, except the $6\sigma$-separation that can hold in any number of dimensions. The validity of the presented analysis in the multidimensional case is corroborated by the toy dataset example in Section \ref{toy_dataset}.

\subsection{Numerical Simulations and Analysis}
\label{numerical_simulations}
In this section, we present numerical simulations of the probabilistic error bounds in \eqref{eq:S_i_bar_proof1} and \eqref{eq:S_i_bar_proof2}. Our goal is to gain insight into: (a) why the method works; (b) the effect of the design choice ($K$) and dataset properties ($\eta, M, T$) on the performance of the model; and (c) analyze possible failure cases. More specifically, we: (a) set $\tau=\tau^*$ such that $P(\Bar{S_i} \geq \tau^* \given p_i = 0) = P(\Bar{S_i} \leq \tau^* \given p_i = 1)$, i.e., $\tau^* = \sqrt{\mathcal{M}_{\Bar{S}_i \given p_i = 0}(t) \cdot \mathcal{M}_{\Bar{S}_i \given p_i = 1}(-t)}$; (b) sweep a single parameter at a time, while the rest are fixed; and (c) optimize for $t$ over $[1, 100]$.
\begin{figure*}[t]
    \centering    
    \includegraphics[width=\linewidth]{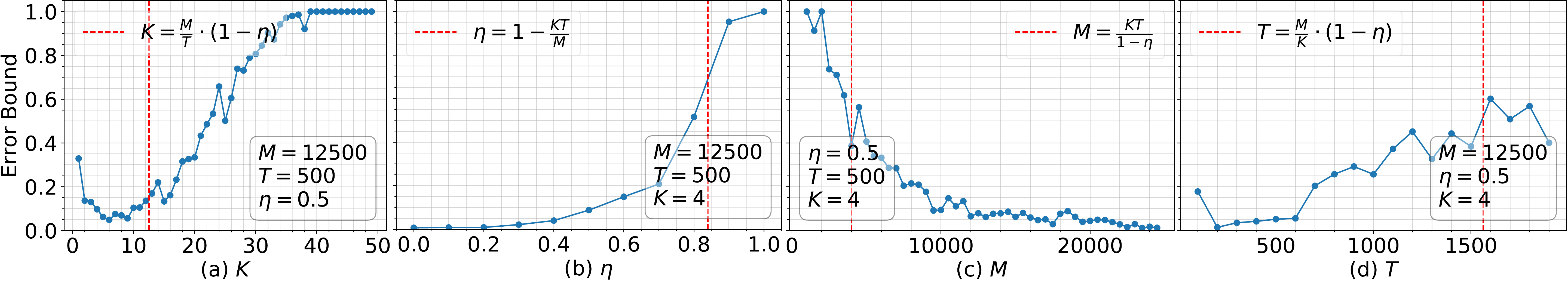}
    \caption{\textbf{Numerical simulations of our probabilistic error upper bound}. From left to right: sweep over $K, \eta, M$ and $T$. We mark with a red dashed vertical line where $K = \frac{M}{T} \cdot (1-\eta)$ holds.}
    \label{fig:numerical_simulations} 
\end{figure*}

\textbf{Discussion.} In Fig. \ref{fig:numerical_simulations}a we study the effect of $K$. As expected, increasing $K$ decreases the error bound initially and from a certain value ($\triangleq K_0$), the error bound increases. Not surprisingly, $K_0 \approx \frac{M}{T} \cdot (1-\eta)$, which is the average number of correctly associated pairs per concept. Throughout Figures \ref{fig:numerical_simulations}a, \ref{fig:numerical_simulations}b, \ref{fig:numerical_simulations}c, \ref{fig:numerical_simulations}d we see the error bound is influenced greatly by this equality, i.e., when $K > K_0$, the model performs well and when $K \leq K_0$, it starts to fail. Specifically, in Fig. \ref{fig:numerical_simulations}c we show that the error bound goes to zero as the size of the dataset ($M$) increases (another point of view is that $K_0$ is increased). For this reason we mark the point where $KT = M (1-\eta)$ by a red dashed line in Fig. \ref{fig:numerical_simulations}. It is clear that for real-world data, concepts are usually not equi-probable and thus assigning a global value for $K$ is sub-optimal. However, this finding allows us to better understand such failure cases and thus choose a reasonable $K$ value. An additional instance where the method fails, gives us insight into why usually the method succeeds. The method will fail for a small number of concepts ($T$) regardless of $K_0$, because for a small number of concepts there is a higher chance that two or more {\it wrongly associated} pairs belong to the same pair of concepts in both modalities. Fortunately, in real-world data $T$ is almost always large, and additionally as $T$ increases, this problem is alleviated by a factor of $\mathcal{O}(T^2)$ (see Appendix \ref{proof_of_theorem}, Eq. \eqref{eq:pr_m_i_tilde}). A simulation of this case is presented in Appendix \ref{low_t_failure_case}. 

\begin{figure*}[ht]
    \centering
    \begin{subfigure}[t]{0.3\textwidth}
        \centering    
        \includegraphics[width=\linewidth]{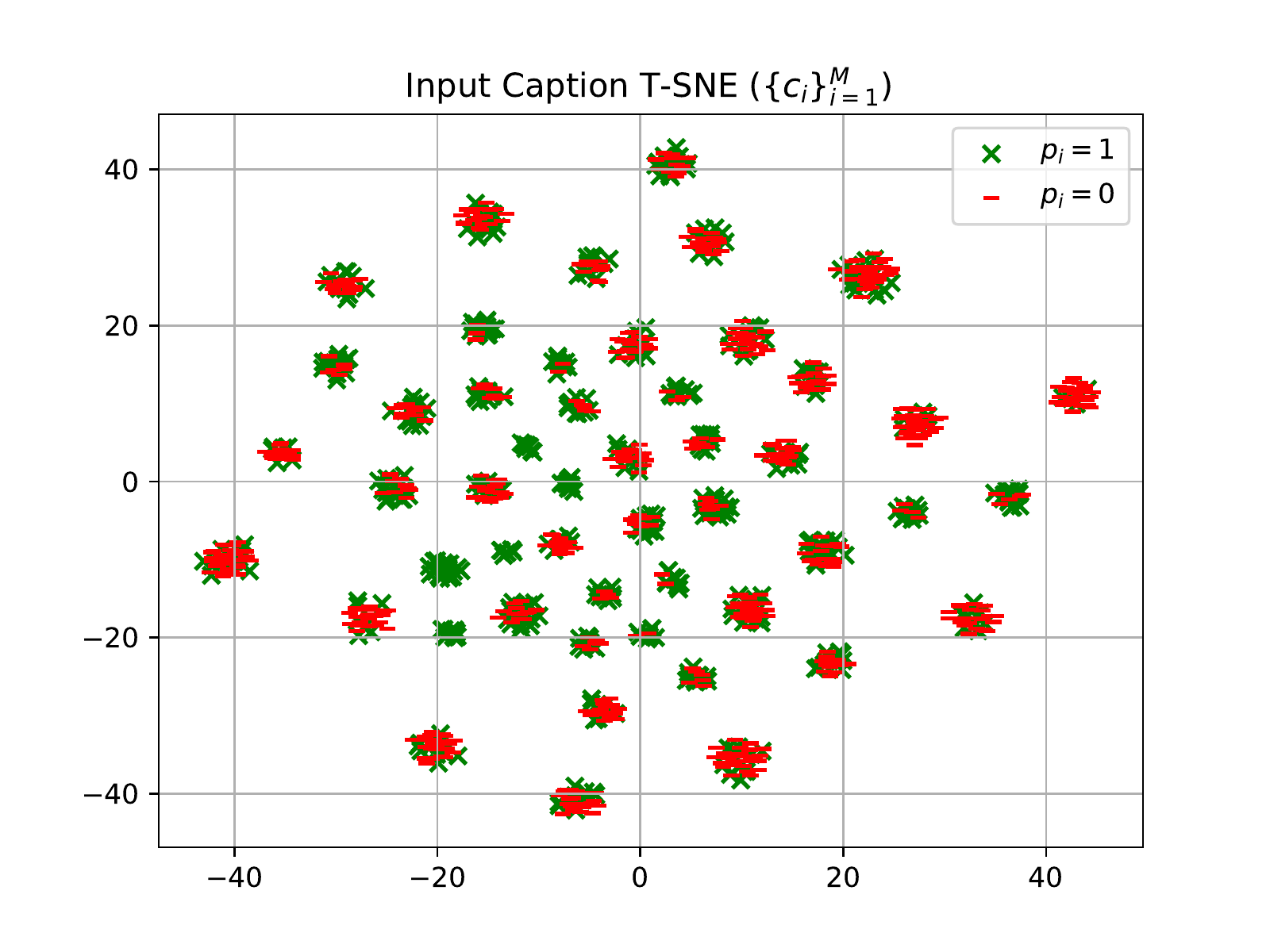}
        \caption{Caption T-SNE}
        \label{fig:toy_caption_tsne} 
    \end{subfigure}%
    \hfill
    \begin{subfigure}[t]{0.3\textwidth}
        \centering
        \includegraphics[width=\linewidth]{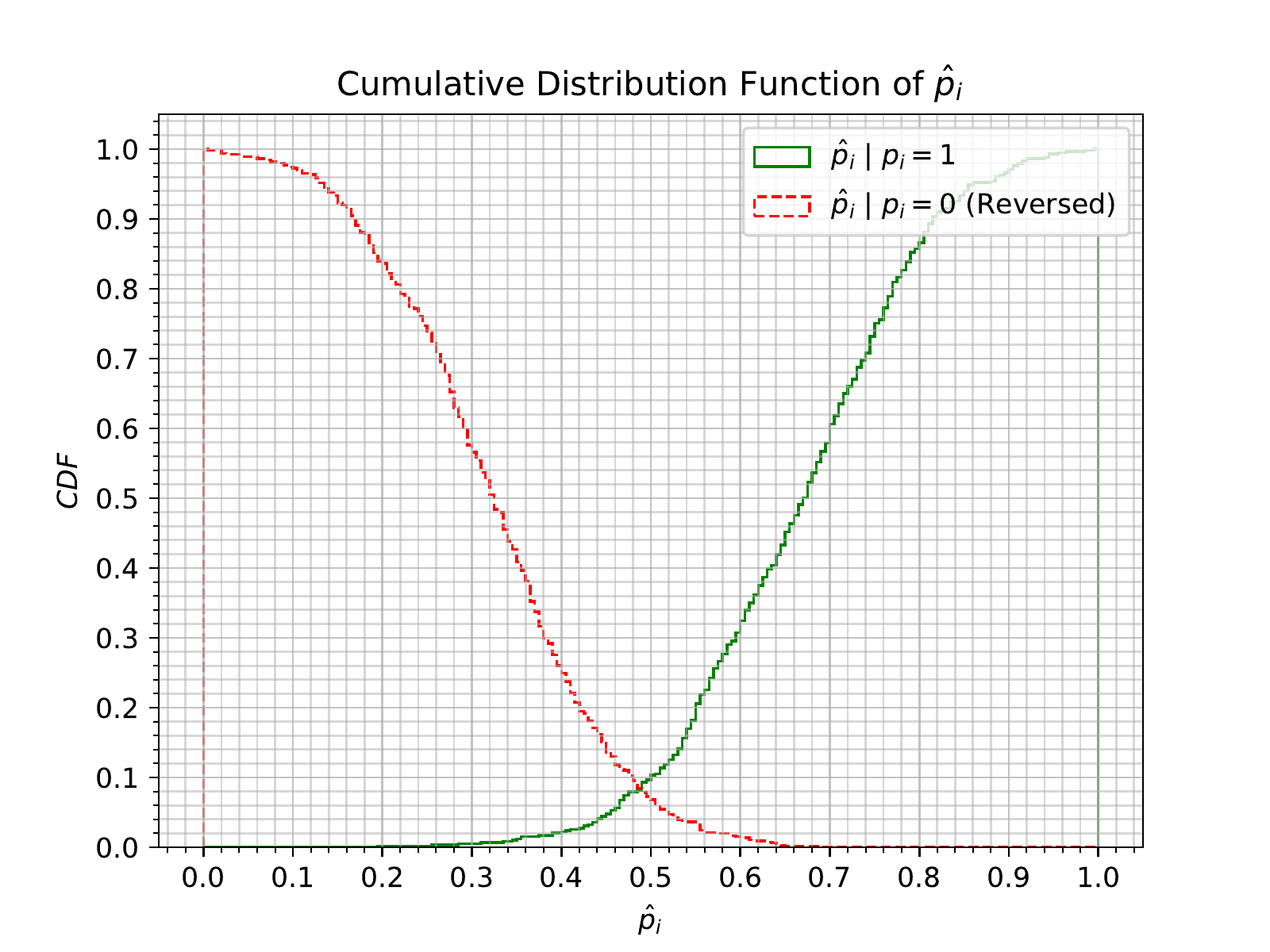}
        \caption{CDF of $\mathbf{\hat{p}}$ (Eq. \eqref{eq:p_i})}
        \label{fig:toy_p_hat}
    \end{subfigure}
    \hfill
    \begin{subfigure}[t]{0.3\textwidth}
        \centering
        \includegraphics[width=\linewidth]{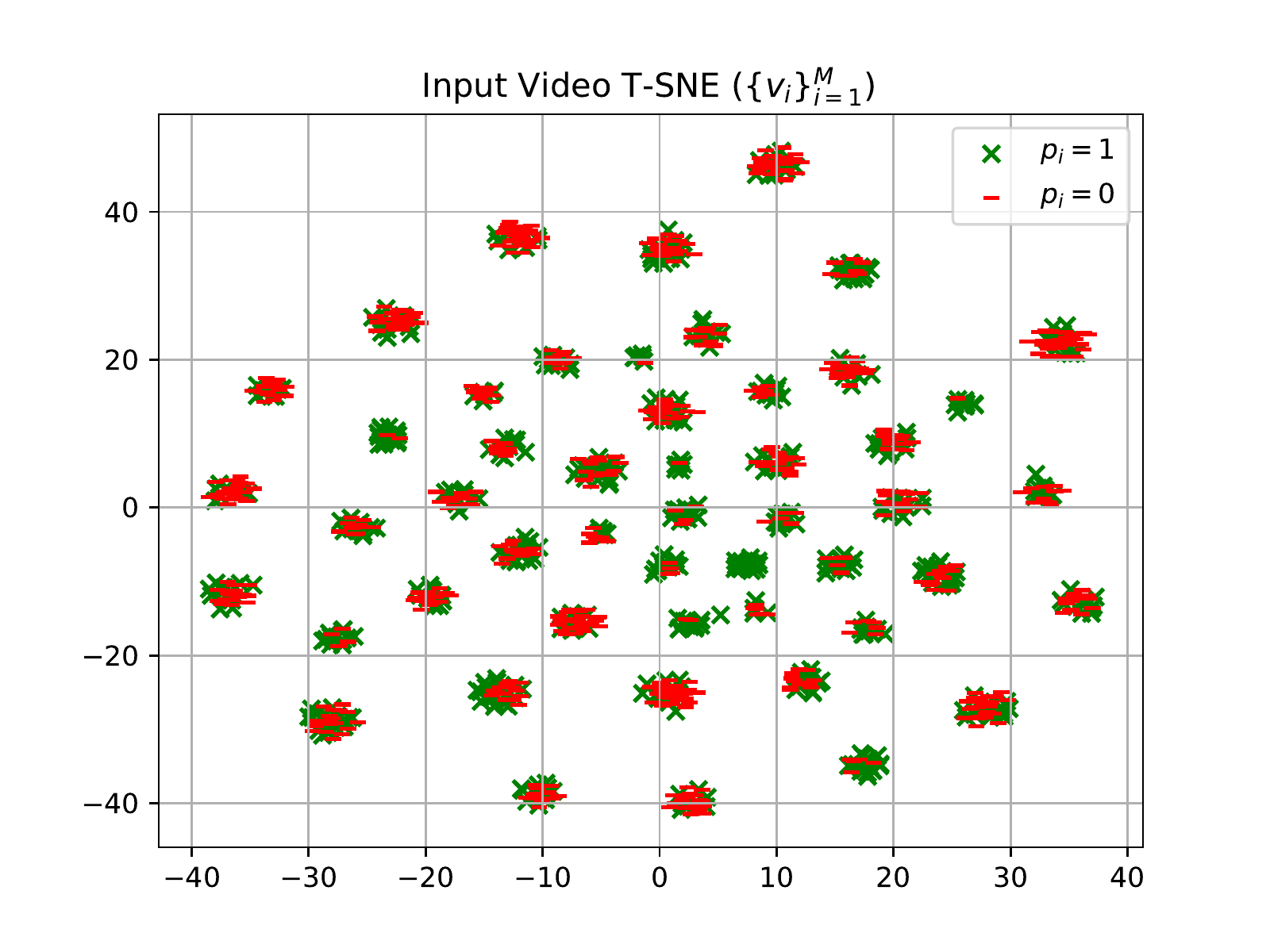}
        \caption{Video T-SNE}
        \label{fig:toy_video_tsne}
    \end{subfigure}
    \caption{\textbf{Toy dataset Visualizations.} (a, c): T-SNE of input embeddings. As we can see, separating between samples with $p_i = 1$ and samples with $p_i=0$ is non-trivial. (b): Cumulative Distribution Function of $\mathbf{\hat{p}}_i$ of toy dataset. The green solid line is the CDF of $\hat{p}_i | p_i = 1$, while the red dashed line is the inverse CDF of $\hat{p}_i | p_i = 0$. Assuming a binary prediction is made based on a hard threshold, it is possible to extract the precision and recall for each threshold from the figure above. For example, for the threshold $0.48$, both precision and recall are $\simeq 0.9$. Best viewed in color.}
\end{figure*}

\section{Experimental Settings}
\subsection{Implementation Details}
\label{implementation_details}

\textbf{Model}. For a fair comparison to the baseline model HTM \cite{miech2019howto100m}, we use the same class of non-linear embedding functions: $f(\bm{v}) = (W_1^v \bm{v} + b_1^v) \circ \sigma(W_2^v(W_1^v \bm{v} + b_1^v) + b_2^v)$, $g(\bm{c}) = (W_1^c \bm{c} + b_1^c) \circ \sigma(W_2^c(W_1^c \bm{c} + b_1^c) + b_2^c)$, where $W_1^v \in \mathbb{R}^{d \times d_{\mathrm{v}}}$, $W_1^c \in \mathbb{R}^{d \times d_{\mathrm{c}}}$, $W_2^v, W_2^c \in \mathbb{R}^{d \times d}$, $b_1^v, b_2^v, b_1^c, b_2^c \in \mathbb{R}^d$ are the learnable parameters, $\sigma$ is an element-wise sigmoid activation and $\circ$ is the element-wise multiplication. We use $d_{\mathrm{v}} = 4096$, $d_{\mathrm{c}} = 300$, and $d = 6144$.

\textbf{Training dataset.} We train our model using the HowTo100M \cite{miech2019howto100m} narrated video dataset that consists of more than 1.2M videos accompanied with automatically generated speech transcription. Similarly to \cite{miech2019howto100m}, we use the provided transcription to create pairs of clip-caption defined by each caption time stamp, where each video clip shorter than 5 seconds is extended symmetrically in time so that the duration is at least 5 seconds. Note that we only use 1.16M videos since some of the videos are no longer available for download.

\textbf{Input caption features.} For the word representations, we use the standard GoogleNews pre-trained word2vec embedding model \cite{mikolov2013distributed}. For the input sentence representations used in Section \ref{mmde} we simply average word representation over each sentence.  

\textbf{Input visual features.} We extract 2D features using ImageNet pre-trained Resnet-152 \cite{he2016deep} at a rate of 1 frame per second. We extract 3D features using Kinetics \cite{carreira2017quo} pre-trained ResNeXt-101 16-frames \cite{hara2018can} at a rate of 24 frames per second. After temporal max pooling we concatenate 2D and 3D features to form a single feature vector per video clip.

\textbf{Loss \& Optimization.} We train our model using the Soft Max Margin loss function described in Section \ref{subsection:soft_max_margin}. We use the ADAM \cite{kingma2015adam} optimizer with a fixed learning rate of $10^{-3}$.

\textbf{Time complexity.} Using FAISS \cite{johnson2019billion}, computation of the Multimodal Density Estimation described in Section \ref{mmde} is done in less than 15 hours over 10 CPUs. Training the model on the large HowTo100M dataset is done on a single V100 GPU and takes less than 24 hours.

Additional implementation details are included in Appendix \ref{additional_imp_details}.

\subsection{Downstream Tasks}
\label{ds_tasks}
\textbf{Video Visual Question Answering (VQA).} The Video VQA task comprises answering questions about videos presented in natural language \cite{antol2015vqa}. Essentially, an instance of VQA includes an input video and a free-form textual query regarding the content in the video, and an expected textual answer. To accommodate this task we fine-tune our learned multimodal representations and evaluate our model on two datasets: MSRVTT-QA and MSVD-QA \cite{xu2017video}. These datasets are based on existing video description datasets. See Table \ref{table:vqa_stats} in Appendix \ref{additional_imp_details} for detailed statistics of each dataset.

Most VQA models use a video and question as input, and the answer is presented as the output of an LSTM unit \cite{hochreiter1997long} or a softmax layer over a set of predetermined answers. However, these types of architectures do not fully utilize the information which exists in coordinated representations, i.e., the representation of the correct answer might likely be closely embedded to the visual representation, given the question. To better utilize our learned multimodal representations specifically for the VQA task, we use a similar architecture to \cite{hu2018learning}, but for video. We learn two more sets of embeddings on top of the pre-trained embeddings that were learned with the HowTo100M dataset: a question+video embedding, i.e., we embed a concatenation of the question and video to a single feature vector; and an answer embedding. We train the model with a max margin ranking loss function to embed an answer close to its question+video. Inference is performed simply with a nearest neighbor search over the set of predetermined answers in the joint video+question and answer space. This model is very simple compared to most VQA models, yet as we show in Table \ref{table:vqa} it is very powerful when built on effective self-supervised pre-trained joint embeddings.

\textbf{Text-To-Video Retrieval.} Text-To-Video Retrieval includes retrieval of video clips based on textual description \cite{Liu19a,mithun2018learning,song2019polysemous}. With a learned joint representation space, retrieval is performed with a nearest neighbor search over the joint embedding space. To evaluate our model we use three different datasets: MSRVTT, MSVD and LSMDC \cite{xu2016msr,chen2011collecting,rohrbach2015dataset}. We use the standard evaluation metrics: recall at $K$ (R@K) for $K = 1, 5, 10$ and median recall (MR). See Table \ref{table:retrieval_stats} in Appendix \ref{additional_imp_details} for detailed statistics of each dataset.

\section{Experiments and Analysis}
\label{experiments}
A toy dataset illustrative results are presented in Section \ref{toy_dataset}. Comparison to ablative baselines and state-of-the-art models is presented in Section \ref{ablative_and_sota}. A design choice analysis is presented in Appendix \ref{design_choice_analysis}. Qualitative examples are presented in Appendix \ref{qualitative_examples}.

\subsection{Multidimensional Toy Dataset}
\label{toy_dataset}
In this section, we demonstrate the effectiveness of our method visually using a toy (synthesized) dataset of mixture of Gaussians of $T$ components in each modality, $\{\mathcal{N}(\bm{\mu}^c_t, \bm{\Sigma}^c_t)\}_{t=1}^T$ and $\{\mathcal{N}(\bm{\mu}^v_t, \bm{\Sigma}^v_t)\}_{k=1}^T$ for caption and video, respectively. {\it Correctly associated} pairs are represented by $v_i \sim \mathcal{N}(\bm{\mu}^v_t, \bm{\Sigma}^v_t), c_i \sim \mathcal{N}(\bm{\mu}^c_t, \bm{\Sigma}^c_t)$, such that $t \in [T], i \in [M]$. {\it Wrongly associated} pairs are represented by $v_i \sim \mathcal{N}(\bm{\mu}^v_m, \bm{\Sigma}^v_m), c_i \sim \mathcal{N}(\bm{\mu}^c_n, \bm{\Sigma}^c_n)$, such that  $m \neq n, \{m, n\} \in [T], i \in [M]$. $\eta \in [0, 1]$ is the noise ratio, such that a {\it wrongly associated} pair is sampled with probability $\eta$, and a {\it correctly associated} pair is sampled with probability $1 - \eta$.

In our experiment, $v_i \in \mathcal{R}^{d_{\mathrm{v}}}, c_i \in \mathcal{R}^{d_{\mathrm{c}}}$, $d_{\mathrm{v}} = 128, d_{\mathrm{c}} = 128$; $\bm{\mu}^c_t \in \mathcal{R}^{d_{\mathrm{c}}}, \bm{\mu}^v_t \in \mathcal{R}^{d_{\mathrm{v}}}, \forall t \in [T]$ are sampled from a uniform multivariate distribution $U_{d_{\mathrm{c}}}(0, 1), U_{d_{\mathrm{v}}}(0, 1)$, respectively for caption and video; $\bm{\Sigma}^c_t \in \mathbb{R}^{d_{\mathrm{c}} \times d_{\mathrm{c}}}, \bm{\Sigma}^v_t \in \mathbb{R}^{d_{\mathrm{v}} \times d_{\mathrm{v}}}$ are diagonal matrices where the diagonals are sampled from a multivariate uniform distribution, $U_{d_{\mathrm{c}}}(0, 0.3), U_{d_{\mathrm{v}}}(0, 0.3)$, respectively for caption and video; $\eta = 0.5, M = 1250, T = 50$; $k = 4$ ($k$-NN parameter).

In Figures \ref{fig:toy_caption_tsne} and \ref{fig:toy_video_tsne} we visualize T-SNE graphs for caption and video embedding spaces, respectively. In Fig. \ref{fig:toy_p_hat} we visualize the empirical cumulative distribution function of $\hat{p}_i$ \eqref{eq:p_i}, the estimated probability of being {\it correctly associated}. Additionally, in Appendix \ref{toy_dataset_appedix} Fig. \ref{fig:empirical_exp}  we empirically reproduce the theoretical graphs in Fig. \ref{fig:numerical_simulations} and Fig. \ref{fig:low_t_failure_case} using the multidimensional toy dataset. These results corroborate the validity of the analysis presented in Thm. \ref{theorem:error_bound} in the multidimensional case. 
\subsection{Ablative Baselines and SOTA Models}
\label{ablative_and_sota}
\begin{table*}[ht]
\small
  \centering
  \setlength\tabcolsep{3.5pt}
  \begin{tabular}{lccccrccccrcccc}
    \toprule
          & \multicolumn{4}{c}{MSRVTT} && \multicolumn{4}{c}{LSMDC} && \multicolumn{4}{c}{MSVD} \\ 
    \cmidrule{2-5} \cmidrule{7-10} \cmidrule{12-15}
    Method & R@1 & R@5 & R@10 & MR && R@1 & R@5 & R@10 & MR && R@1 & R@5 & R@10 & MR \\ 
    \midrule
    \multicolumn{15}{c}{\underline{\textbf{Zero-Shot}}}\\
    Random & 0.1 & 0.5 & 1.0 & 500.0 && 0.1 & 0.5 & 1.0 & 500.0 && 0.15 & 0.75 & 1.49 & 335 \\
    MIL-NCE$^\dagger$ \cite{miech2020end} & \textbf{9.9} & \textbf{24.0} & \textbf{32.4} & \textbf{29.5} && -- & -- & -- & -- && -- & -- & -- & -- \\ 
    \midrule
    HTM-PT$^*$ \cite{miech2019howto100m} & 7.5 & 21.2 & \textbf{29.6} & 38.0 && 4.0 & 9.8 & 14.0 & 137.0 && 12.86 & 33.06 & 45.83 & 13.0 \\
    Ours & \textbf{8.0} & \textbf{21.3} & 29.3 & \textbf{33.0} && \textbf{4.2} & \textbf{11.6} & \textbf{17.1} & \textbf{119.0} && \textbf{13.66} & \textbf{35.7} & \textbf{47.74} & \textbf{12.0} \\
    \midrule
    \midrule
    \multicolumn{15}{c}{\underline{\textbf{Fine-Tuned}}}\\
    CE$^\ddagger$ \cite{Liu19a} & 18.2 & 46.0 & 60.7 & 7.0 && 11.2 & 26.9 & 34.8 & 25.0 && 19.8 & 49.0 & 63.8 & 6.0 \\
    \midrule
    JEMC \cite{mithun2018learning} & 7.0 & 20.9 & 29.7 & 38.0 && -- & -- & -- & -- && \textbf{20.3} & 47.8 & 61.1 & 6.0 \\
    JSFusion \cite{yu2018joint} & 10.2 & 31.2 & 43.2 & 13.0 && 9.1 & 21.2 & 34.1 & 36 && -- & -- & -- & -- \\
    MoEE \cite{miech2018learning} & 14.2 & 39.2 & \textbf{53.8} & 9 && \textbf{10.1} & \textbf{25.6} & \textbf{34.6} & \textbf{27} && -- & -- & -- & -- \\
    \midrule
    HTM-no-PT \cite{miech2019howto100m} & 12.4 & 36.0 & 52.0 & 10.0 && 5.8 & 18.8 & 28.4 & 45.0 && 13.0 & 37.43 & 52.41 & 10.0 \\
    HTM-PT$^*$ \cite{miech2019howto100m} & 14.9 & 40.2 & 52.8 & 9.0 && \textbf{7.1} & 19.6 & 27.9 & 40.0 && 15.52 & 40.93 & 55.7 & 8.0 \\
    Ours & \textbf{17.4} & \textbf{41.6} & \textbf{53.6} & \textbf{8.0} && 6.4 & \textbf{19.8} & \textbf{28.4} & \textbf{39.0} && \textbf{20.3} & \textbf{48.97} & \textbf{63.26} & \textbf{6.0} \\
    \bottomrule
  \end{tabular}
  \caption{Text-To-Video retrieval. Zero-Shot: training was done only with HowTo100M dataset. Fine-Tuned: model was fine-tuned with the relevant benchmark dataset. For MR the lower the better. *: results for MSRVTT and LSMDC are from \cite{miech2019howto100m}, while results for MSVD have been reproduced. $^\dagger$: MIL-NCE \cite{miech2020end} use additional clip-caption pairs. $^\ddagger$: CE \cite{Liu19a} use extra labeled data in the form of pre-trained semantic embeddings which include ‘general’ features such as motion, appearance, scene features and OCR.}
  \label{table:text_to_video}
\end{table*}
\begin{table}
\small
  \centering
  \setlength\tabcolsep{3pt}
  \begin{tabular}{lcc}
    \toprule
    & MSRVTT-QA [\%] & MSVD-QA [\%] \\ 
    \midrule
    ST-VQA \shortcite{jang2017tgif} & 30.09 & 31.3 \\ 
    Co-Mem \shortcite{gao2018motion} & 32.0 & 31.7 \\ 
    AMU \shortcite{xu2017video} & 32.5 & 32.0 \\
    HMEMA \shortcite{fan2019heterogeneous} & 33.0 & 33.7\\
    \midrule
    HTM-no-PT \shortcite{miech2019howto100m} & 27.05 & 33.8\\
    HTM-PT \shortcite{miech2019howto100m} & 34.38 & 34.83\\
    Ours & \textbf{35.06} & \textbf{35.13}\\ 
    \bottomrule
  \end{tabular}
  \caption{Video Question Answering. Results of ST-VQA and Co-Mem taken from \cite{fan2019heterogeneous}}
  \label{table:vqa}
\end{table}
We compare our proposed model against two ablative baselines and multiple task-specific state-of-the-art (SOTA) models:

\textbf{HTM-PT \cite{miech2019howto100m}}. The model (architecture and loss function) used in \cite{miech2019howto100m}. This baseline is pre-trained (PT) on the HowTo100M dataset. It is the exact architecture described in Section \ref{implementation_details} for our model. The only differentiating element is the loss function. \cite{miech2019howto100m} use the max margin ranking loss function, while we use our proposed Soft Max Margin ranking loss function. Since the model is also trained identically to our own model it is clear that any gain in performance over this baseline is due to our novel noise estimation based density estimation component.

\textbf{HTM-no-PT \cite{miech2019howto100m}}. The same model from above, but without pre-training (no-PT) on the HowTo100M dataset. The (under) performance of this baseline on downstream tasks demonstrates the potential gain of utilizing self-supervised speech-visual correspondence training.

\textbf{Task specific state-of-the-art models}. After fine-tuning for downstream tasks we compare our proposed model to state-of-the-art models for each task and each dataset. \cite{gao2018motion,xu2017video,fan2019heterogeneous,jang2017tgif} for VQA, and \cite{Liu19a,mithun2018learning,yu2018joint,miech2018learning,miech2019howto100m,miech2020end} for Text-To-Video Retrieval.

Tables \ref{table:text_to_video} and \ref{table:vqa} show the result for Text-To-Video Retrieval and Video Question Answering, respectively. Table \ref{table:unfair_evaluations} in Appendix \ref{unfair_evaluations} shows the results for Zero-Shot Text-To-Video Retrieval under `unfair' settings. We summarize key insights below:
\begin{itemize}[topsep=0pt, leftmargin=*]
    \item[--] Our model consistently outperforms the baselines (HTM-PT, HTM-no-PT \cite{miech2019howto100m}) in both Visual Question Answering and Text-To-Video Retrieval on five different datsets. 
    \item[--] We set a new state-of-the-art performance for two Visual Question Answering datasets: MSRVTT-QA and MSVD-QA.
    \item[--] We set a new state-of-the-art performance for Zero-Shot Text-To-Video Retrieval on two datasets: LSMDC and MSVD. For MSRVTT, we hypothesize that MIL-NCE \cite{miech2020end} performs better due to two reasons: i) they train end-to-end, unlike our baseline model, which operates on pre-proccessed embeddings; and ii) they utilize additional clip-caption pairs by associating a clip with its adjacent (in time) captions. In fact, when those additional pairs are not used our model outperforms theirs.
    \item[--] We set a new state-of-the-art performance for (fine-tuned) Text-To-Video Retrieval on two datasets: MSRVTT and MSVD.
    \item[--] We demonstrate that our model outperforms or is at least on par with the performance of HTM-PT \cite{miech2019howto100m} even given a setting which is a clear disadvantage such as training it without 3D features (i.e., only 2D). See Table \ref{table:unfair_evaluations} in Appendix \ref{unfair_evaluations}. This shows:
    \begin{enumerate}[label=(\roman*)]
        \item The power of our noise estimation method and its potential.
        \item Integrating our multimodal density estimation component allows saving time and/or computation power by training and running inference with only 2D features, without (or with minor) performance degradation.
    \end{enumerate}
\end{itemize}

\section{Summary}
In this work, we showed that the problem of noise estimation in multimodal data can be effectively reduced to a multimodal density estimation task. Based on this efficient noise estimation we proposed a novel building block for noise robust multimodal representation learning that can be integrated into many multimodal learning models and improve their performance instantly. We demonstrated how to integrate our building block into the max margin ranking loss function (Soft Max Margin) and it can similarly be integrated into various architectures and losses. We trained Soft Max Margin on the self-supervised proxy task of speech-visual correspondence that is known to be highly noisy. We further evaluated Soft Max Margin on two different downstream tasks: Visual Question Answering and Text-to-Video Retrieval; and achieved comparable state-of-the-art performance on five different datasets. For supporting the empirical results and analyzing failure cases, we provided a theoretical probabilistic error bound. These results emphasize the importance of self-supervised multimodal representation learning for advancing the state of the art in challenging multimodal artificial intelligence tasks. 
\small{\bibliography{ref.bib}}

\begin{thebibliography}{69}
\providecommand{\natexlab}[1]{#1}
\providecommand{\url}[1]{\texttt{#1}}
\providecommand{\urlprefix}{URL }
\expandafter\ifx\csname urlstyle\endcsname\relax
  \providecommand{\doi}[1]{doi:\discretionary{}{}{}#1}\else
  \providecommand{\doi}{doi:\discretionary{}{}{}\begingroup
  \urlstyle{rm}\Url}\fi

\bibitem[{Amrani et~al.(2019)Amrani, Ben-Ari, Hakim, and
  Bronstein}]{amrani2019learning}
Amrani, E.; Ben-Ari, R.; Hakim, T.; and Bronstein, A. 2019.
\newblock Learning to Detect and Retrieve Objects From Unlabeled Videos.
\newblock In \emph{2019 IEEE/CVF International Conference on Computer Vision
  Workshop (ICCVW)}, 3713--3717. IEEE.

\bibitem[{Antol et~al.(2015)Antol, Agrawal, Lu, Mitchell, Batra,
  Lawrence~Zitnick, and Parikh}]{antol2015vqa}
Antol, S.; Agrawal, A.; Lu, J.; Mitchell, M.; Batra, D.; Lawrence~Zitnick, C.;
  and Parikh, D. 2015.
\newblock Vqa: Visual question answering.
\newblock In \emph{Proceedings of the IEEE international conference on computer
  vision}, 2425--2433.

\bibitem[{Arandjelovic and Zisserman(2017)}]{arandjelovic2017look}
Arandjelovic, R.; and Zisserman, A. 2017.
\newblock Look, listen and learn.
\newblock In \emph{Proceedings of the IEEE International Conference on Computer
  Vision}, 609--617.

\bibitem[{Carreira and Zisserman(2017)}]{carreira2017quo}
Carreira, J.; and Zisserman, A. 2017.
\newblock Quo vadis, action recognition? a new model and the kinetics dataset.
\newblock In \emph{proceedings of the IEEE Conference on Computer Vision and
  Pattern Recognition}, 6299--6308.

\bibitem[{Chen and Dolan(2011)}]{chen2011collecting}
Chen, D.~L.; and Dolan, W.~B. 2011.
\newblock Collecting highly parallel data for paraphrase evaluation.
\newblock In \emph{Proceedings of the 49th Annual Meeting of the Association
  for Computational Linguistics: Human Language Technologies-Volume 1},
  190--200. Association for Computational Linguistics.

\bibitem[{Fan et~al.(2019)Fan, Zhang, Zhang, Wang, Zhang, and
  Huang}]{fan2019heterogeneous}
Fan, C.; Zhang, X.; Zhang, S.; Wang, W.; Zhang, C.; and Huang, H. 2019.
\newblock Heterogeneous memory enhanced multimodal attention model for video
  question answering.
\newblock In \emph{Proceedings of the IEEE Conference on Computer Vision and
  Pattern Recognition}, 1999--2007.

\bibitem[{Frome et~al.(2013)Frome, Corrado, Shlens, Bengio, Dean, Ranzato, and
  Mikolov}]{frome2013devise}
Frome, A.; Corrado, G.~S.; Shlens, J.; Bengio, S.; Dean, J.; Ranzato, M.; and
  Mikolov, T. 2013.
\newblock Devise: A deep visual-semantic embedding model.
\newblock In \emph{Advances in neural information processing systems},
  2121--2129.

\bibitem[{Gao et~al.(2018)Gao, Ge, Chen, and Nevatia}]{gao2018motion}
Gao, J.; Ge, R.; Chen, K.; and Nevatia, R. 2018.
\newblock Motion-appearance co-memory networks for video question answering.
\newblock In \emph{Proceedings of the IEEE Conference on Computer Vision and
  Pattern Recognition}, 6576--6585.

\bibitem[{Ghosh, Kumar, and Sastry(2017)}]{ghosh2017robust}
Ghosh, A.; Kumar, H.; and Sastry, P. 2017.
\newblock Robust loss functions under label noise for deep neural networks.
\newblock In \emph{Thirty-First AAAI Conference on Artificial Intelligence}.

\bibitem[{Goldberger and Ben-Reuven(2017)}]{Goldberger2017TrainingDN}
Goldberger, J.; and Ben-Reuven, E. 2017.
\newblock Training deep neural-networks using a noise adaptation layer.
\newblock In \emph{ICLR}.

\bibitem[{Hara, Kataoka, and Satoh(2018)}]{hara2018can}
Hara, K.; Kataoka, H.; and Satoh, Y. 2018.
\newblock Can spatiotemporal 3d cnns retrace the history of 2d cnns and
  imagenet?
\newblock In \emph{Proceedings of the IEEE conference on Computer Vision and
  Pattern Recognition}, 6546--6555.

\bibitem[{He et~al.(2016)He, Zhang, Ren, and Sun}]{he2016deep}
He, K.; Zhang, X.; Ren, S.; and Sun, J. 2016.
\newblock Deep residual learning for image recognition.
\newblock In \emph{Proceedings of the IEEE conference on computer vision and
  pattern recognition}, 770--778.

\bibitem[{Hochreiter and Schmidhuber(1997)}]{hochreiter1997long}
Hochreiter, S.; and Schmidhuber, J. 1997.
\newblock Long short-term memory.
\newblock \emph{Neural computation} 9(8): 1735--1780.

\bibitem[{Hu, Chao, and Sha(2018)}]{hu2018learning}
Hu, H.; Chao, W.-L.; and Sha, F. 2018.
\newblock Learning answer embeddings for visual question answering.
\newblock In \emph{Proceedings of the IEEE Conference on Computer Vision and
  Pattern Recognition}, 5428--5436.

\bibitem[{Jang et~al.(2017)Jang, Song, Yu, Kim, and Kim}]{jang2017tgif}
Jang, Y.; Song, Y.; Yu, Y.; Kim, Y.; and Kim, G. 2017.
\newblock Tgif-qa: Toward spatio-temporal reasoning in visual question
  answering.
\newblock In \emph{Proceedings of the IEEE Conference on Computer Vision and
  Pattern Recognition}, 2758--2766.

\bibitem[{Jiang et~al.(2018)Jiang, Zhou, Leung, Li, and
  Fei{-}Fei}]{lu2018mentornet}
Jiang, L.; Zhou, Z.; Leung, T.; Li, L.; and Fei{-}Fei, L. 2018.
\newblock MentorNet: Learning Data-Driven Curriculum for Very Deep Neural
  Networks on Corrupted Labels.
\newblock In \emph{Proceedings of the 35th International Conference on Machine
  Learning, {ICML} 2018, Stockholmsm{\"{a}}ssan, Stockholm, Sweden, July 10-15,
  2018}, volume~80 of \emph{Proceedings of Machine Learning Research},
  2309--2318. {PMLR}.

\bibitem[{Johnson, Douze, and J{\'e}gou(2019)}]{johnson2019billion}
Johnson, J.; Douze, M.; and J{\'e}gou, H. 2019.
\newblock Billion-scale similarity search with GPUs.
\newblock \emph{IEEE Transactions on Big Data} .

\bibitem[{Kingma and Ba(2015)}]{kingma2015adam}
Kingma, D.~P.; and Ba, J. 2015.
\newblock Adam: A method for stochastic optimization.
\newblock In \emph{ICLR 2015}.

\bibitem[{Koller and Friedman(2009)}]{koller2009probabilistic}
Koller, D.; and Friedman, N. 2009.
\newblock \emph{Probabilistic graphical models: principles and techniques}.
\newblock MIT press.

\bibitem[{Korbar, Tran, and Torresani(2018)}]{korbar2018cooperative}
Korbar, B.; Tran, D.; and Torresani, L. 2018.
\newblock Cooperative learning of audio and video models from self-supervised
  synchronization.
\newblock In \emph{Advances in Neural Information Processing Systems},
  7763--7774.

\bibitem[{Li et~al.(2019)Li, Wong, Zhao, and Kankanhalli}]{li2019learning}
Li, J.; Wong, Y.; Zhao, Q.; and Kankanhalli, M.~S. 2019.
\newblock Learning to learn from noisy labeled data.
\newblock In \emph{Proceedings of the IEEE Conference on Computer Vision and
  Pattern Recognition}, 5051--5059.

\bibitem[{Li et~al.(2017)Li, Yang, Song, Cao, Luo, and Li}]{li2017learning}
Li, Y.; Yang, J.; Song, Y.; Cao, L.; Luo, J.; and Li, L.-J. 2017.
\newblock Learning from noisy labels with distillation.
\newblock In \emph{Proceedings of the IEEE International Conference on Computer
  Vision}, 1910--1918.

\bibitem[{Liu et~al.(2019)Liu, Albanie, Nagrani, and Zisserman}]{Liu19a}
Liu, Y.; Albanie, S.; Nagrani, A.; and Zisserman, A. 2019.
\newblock Use What You Have: Video Retrieval Using Representations From
  Collaborative Experts.
\newblock In \emph{British Machine Vision Conference}.

\bibitem[{Mao et~al.(2014)Mao, Xu, Yang, Wang, Huang, and Yuille}]{mao2014deep}
Mao, J.; Xu, W.; Yang, Y.; Wang, J.; Huang, Z.; and Yuille, A. 2014.
\newblock Deep captioning with multimodal recurrent neural networks (m-rnn).
\newblock \emph{arXiv preprint arXiv:1412.6632} .

\bibitem[{McLachlan and Krishnan(2007)}]{mclachlan2007algorithm}
McLachlan, G.~J.; and Krishnan, T. 2007.
\newblock \emph{The EM algorithm and extensions}, volume 382.
\newblock John Wiley \& Sons.

\bibitem[{Miech et~al.(2020)Miech, Alayrac, Smaira, Laptev, Sivic, and
  Zisserman}]{miech2020end}
Miech, A.; Alayrac, J.-B.; Smaira, L.; Laptev, I.; Sivic, J.; and Zisserman, A.
  2020.
\newblock End-to-end learning of visual representations from uncurated
  instructional videos.
\newblock In \emph{Proceedings of the IEEE/CVF Conference on Computer Vision
  and Pattern Recognition}, 9879--9889.

\bibitem[{Miech, Laptev, and Sivic(2018)}]{miech2018learning}
Miech, A.; Laptev, I.; and Sivic, J. 2018.
\newblock Learning a text-video embedding from incomplete and heterogeneous
  data.
\newblock \emph{arXiv preprint arXiv:1804.02516} .

\bibitem[{Miech et~al.(2019)Miech, Zhukov, Alayrac, Tapaswi, Laptev, and
  Sivic}]{miech2019howto100m}
Miech, A.; Zhukov, D.; Alayrac, J.-B.; Tapaswi, M.; Laptev, I.; and Sivic, J.
  2019.
\newblock Howto100M: Learning a text-video embedding by watching hundred
  million narrated video clips.
\newblock In \emph{Proceedings of the IEEE International Conference on Computer
  Vision}, 2630--2640.

\bibitem[{Mikolov et~al.(2013)Mikolov, Sutskever, Chen, Corrado, and
  Dean}]{mikolov2013distributed}
Mikolov, T.; Sutskever, I.; Chen, K.; Corrado, G.~S.; and Dean, J. 2013.
\newblock Distributed representations of words and phrases and their
  compositionality.
\newblock In \emph{Advances in neural information processing systems},
  3111--3119.

\bibitem[{Misra, Zitnick, and Hebert(2016)}]{misra2016shuffle}
Misra, I.; Zitnick, C.~L.; and Hebert, M. 2016.
\newblock Shuffle and learn: unsupervised learning using temporal order
  verification.
\newblock In \emph{European Conference on Computer Vision}, 527--544. Springer.

\bibitem[{Mithun et~al.(2018)Mithun, Li, Metze, and
  Roy-Chowdhury}]{mithun2018learning}
Mithun, N.~C.; Li, J.; Metze, F.; and Roy-Chowdhury, A.~K. 2018.
\newblock Learning joint embedding with multimodal cues for cross-modal
  video-text retrieval.
\newblock In \emph{Proceedings of the 2018 ACM on International Conference on
  Multimedia Retrieval}, 19--27.

\bibitem[{Moriya et~al.(2019)Moriya, Sanabria, Metze, and
  Jones}]{moriya2019grounding}
Moriya, Y.; Sanabria, R.; Metze, F.; and Jones, G.~J. 2019.
\newblock Grounding Object Detections With Transcriptions.
\newblock In \emph{ICML Workshop 2019}.

\bibitem[{Noroozi and Favaro(2016)}]{noroozi2016unsupervised}
Noroozi, M.; and Favaro, P. 2016.
\newblock Unsupervised learning of visual representations by solving jigsaw
  puzzles.
\newblock In \emph{European Conference on Computer Vision}, 69--84. Springer.

\bibitem[{Papamakarios, Pavlakou, and Murray(2017)}]{papamakarios2017masked}
Papamakarios, G.; Pavlakou, T.; and Murray, I. 2017.
\newblock Masked autoregressive flow for density estimation.
\newblock In \emph{Advances in Neural Information Processing Systems},
  2338--2347.

\bibitem[{Pathak et~al.(2016)Pathak, Krahenbuhl, Donahue, Darrell, and
  Efros}]{pathak2016context}
Pathak, D.; Krahenbuhl, P.; Donahue, J.; Darrell, T.; and Efros, A.~A. 2016.
\newblock Context encoders: Feature learning by inpainting.
\newblock In \emph{Proceedings of the IEEE conference on computer vision and
  pattern recognition}, 2536--2544.

\bibitem[{Reed et~al.(2016)Reed, Akata, Yan, Logeswaran, Schiele, and
  Lee}]{reed2016generative}
Reed, S.; Akata, Z.; Yan, X.; Logeswaran, L.; Schiele, B.; and Lee, H. 2016.
\newblock Generative adversarial text to image synthesis.
\newblock \emph{arXiv preprint arXiv:1605.05396} .

\bibitem[{Reed et~al.(2015)Reed, Lee, Anguelov, Szegedy, Erhan, and
  Rabinovich}]{reed2015training}
Reed, S.; Lee, H.; Anguelov, D.; Szegedy, C.; Erhan, D.; and Rabinovich, A.
  2015.
\newblock Training deep neural networks on noisy labels with bootstrapping.
\newblock In \emph{ICLR Workshop 2015}.

\bibitem[{Rohrbach, Rohrbach, and Schiele(2015)}]{rohrbach2015long}
Rohrbach, A.; Rohrbach, M.; and Schiele, B. 2015.
\newblock The long-short story of movie description.
\newblock In \emph{German conference on pattern recognition}, 209--221.
  Springer.

\bibitem[{Rohrbach et~al.(2015)Rohrbach, Rohrbach, Tandon, and
  Schiele}]{rohrbach2015dataset}
Rohrbach, A.; Rohrbach, M.; Tandon, N.; and Schiele, B. 2015.
\newblock A dataset for movie description.
\newblock In \emph{Proceedings of the IEEE conference on computer vision and
  pattern recognition}, 3202--3212.

\bibitem[{Salakhutdinov and Hinton(2009)}]{salakhutdinov2009deep}
Salakhutdinov, R.; and Hinton, G. 2009.
\newblock Deep boltzmann machines.
\newblock In \emph{Artificial intelligence and statistics}, 448--455.

\bibitem[{Schroff, Kalenichenko, and Philbin(2015)}]{schroff2015facenet}
Schroff, F.; Kalenichenko, D.; and Philbin, J. 2015.
\newblock Facenet: A unified embedding for face recognition and clustering.
\newblock In \emph{Proceedings of the IEEE conference on computer vision and
  pattern recognition}, 815--823.

\bibitem[{Silverman(2018)}]{silverman2018density}
Silverman, B.~W. 2018.
\newblock \emph{Density estimation for statistics and data analysis}.
\newblock Routledge.

\bibitem[{Socher et~al.(2014)Socher, Karpathy, Le, Manning, and
  Ng}]{socher2014grounded}
Socher, R.; Karpathy, A.; Le, Q.~V.; Manning, C.~D.; and Ng, A.~Y. 2014.
\newblock Grounded compositional semantics for finding and describing images
  with sentences.
\newblock \emph{Transactions of the Association for Computational Linguistics}
  2: 207--218.

\bibitem[{Song and Soleymani(2019)}]{song2019polysemous}
Song, Y.; and Soleymani, M. 2019.
\newblock Polysemous visual-semantic embedding for cross-modal retrieval.
\newblock In \emph{Proceedings of the IEEE Conference on Computer Vision and
  Pattern Recognition}, 1979--1988.

\bibitem[{Srivastava, Mansimov, and
  Salakhudinov(2015)}]{srivastava2015unsupervised}
Srivastava, N.; Mansimov, E.; and Salakhudinov, R. 2015.
\newblock Unsupervised learning of video representations using lstms.
\newblock In \emph{International conference on machine learning}, 843--852.

\bibitem[{Srivastava and Salakhutdinov(2012)}]{srivastava2012learning}
Srivastava, N.; and Salakhutdinov, R. 2012.
\newblock Learning representations for multimodal data with deep belief nets.
\newblock In \emph{International conference on machine learning workshop},
  volume~79.

\bibitem[{Stone(1994)}]{stone1994use}
Stone, C.~J. 1994.
\newblock The use of polynomial splines and their tensor products in
  multivariate function estimation.
\newblock \emph{The Annals of Statistics} 118--171.

\bibitem[{Sukhbaatar et~al.(2015)Sukhbaatar, Estrach, Paluri, Bourdev, and
  Fergus}]{sukhbaatar2015training}
Sukhbaatar, S.; Estrach, J.~B.; Paluri, M.; Bourdev, L.; and Fergus, R. 2015.
\newblock Training convolutional networks with noisy labels.
\newblock In \emph{ICLR Workshop 2015}.

\bibitem[{Sun et~al.(2019{\natexlab{a}})Sun, Baradel, Murphy, and
  Schmid}]{sun2019contrastive}
Sun, C.; Baradel, F.; Murphy, K.; and Schmid, C. 2019{\natexlab{a}}.
\newblock Contrastive bidirectional transformer for temporal representation
  learning.
\newblock \emph{arXiv preprint arXiv:1906.05743} .

\bibitem[{Sun et~al.(2019{\natexlab{b}})Sun, Myers, Vondrick, Murphy, and
  Schmid}]{sun2019videobert}
Sun, C.; Myers, A.; Vondrick, C.; Murphy, K.; and Schmid, C.
  2019{\natexlab{b}}.
\newblock Videobert: A joint model for video and language representation
  learning.
\newblock In \emph{Proceedings of the IEEE International Conference on Computer
  Vision}, 7464--7473.

\bibitem[{Tanaka et~al.(2018)Tanaka, Ikami, Yamasaki, and
  Aizawa}]{tanaka2018joint}
Tanaka, D.; Ikami, D.; Yamasaki, T.; and Aizawa, K. 2018.
\newblock Joint optimization framework for learning with noisy labels.
\newblock In \emph{Proceedings of the IEEE Conference on Computer Vision and
  Pattern Recognition}, 5552--5560.

\bibitem[{Terrell and Scott(1992)}]{terrell1992variable}
Terrell, G.~R.; and Scott, D.~W. 1992.
\newblock Variable kernel density estimation.
\newblock \emph{The Annals of Statistics} 1236--1265.

\bibitem[{Tian, Krishnan, and Isola(2019)}]{tian2019contrastive}
Tian, Y.; Krishnan, D.; and Isola, P. 2019.
\newblock Contrastive multiview coding.
\newblock \emph{arXiv preprint arXiv:1906.05849} .

\bibitem[{Uria, Murray, and Larochelle(2014)}]{uria2014deep}
Uria, B.; Murray, I.; and Larochelle, H. 2014.
\newblock A deep and tractable density estimator.
\newblock In \emph{International Conference on Machine Learning}, 467--475.

\bibitem[{Van~Rooyen, Menon, and Williamson(2015)}]{van2015learning}
Van~Rooyen, B.; Menon, A.; and Williamson, R.~C. 2015.
\newblock Learning with symmetric label noise: The importance of being
  unhinged.
\newblock In \emph{Advances in Neural Information Processing Systems}, 10--18.

\bibitem[{Vendrov et~al.(2016)Vendrov, Kiros, Fidler, and
  Urtasun}]{ivan2016order}
Vendrov, I.; Kiros, R.; Fidler, S.; and Urtasun, R. 2016.
\newblock Order-Embeddings of Images and Language.
\newblock In Bengio, Y.; and LeCun, Y., eds., \emph{4th International
  Conference on Learning Representations, {ICLR} 2016, San Juan, Puerto Rico,
  May 2-4, 2016, Conference Track Proceedings}.

\bibitem[{Venugopalan et~al.(2015)Venugopalan, Xu, Donahue, Rohrbach, Mooney,
  and Saenko}]{venugopalan2015translating}
Venugopalan, S.; Xu, H.; Donahue, J.; Rohrbach, M.; Mooney, R.; and Saenko, K.
  2015.
\newblock Translating Videos to Natural Language Using Deep Recurrent Neural
  Networks.
\newblock In \emph{Proceedings of the 2015 Conference of the North American
  Chapter of the Association for Computational Linguistics: Human Language
  Technologies}, 1494--1504.

\bibitem[{Vondrick, Pirsiavash, and Torralba(2016)}]{vondrick2016generating}
Vondrick, C.; Pirsiavash, H.; and Torralba, A. 2016.
\newblock Generating videos with scene dynamics.
\newblock In \emph{Advances in neural information processing systems},
  613--621.

\bibitem[{Wang et~al.(2014)Wang, Song, Leung, Rosenberg, Wang, Philbin, Chen,
  and Wu}]{wang2014learning}
Wang, J.; Song, Y.; Leung, T.; Rosenberg, C.; Wang, J.; Philbin, J.; Chen, B.;
  and Wu, Y. 2014.
\newblock Learning fine-grained image similarity with deep ranking.
\newblock In \emph{Proceedings of the IEEE Conference on Computer Vision and
  Pattern Recognition}, 1386--1393.

\bibitem[{Wang, Li, and Lazebnik(2016)}]{wang2016learning}
Wang, L.; Li, Y.; and Lazebnik, S. 2016.
\newblock Learning deep structure-preserving image-text embeddings.
\newblock In \emph{Proceedings of the IEEE conference on computer vision and
  pattern recognition}, 5005--5013.

\bibitem[{Wang and Wang(2015)}]{wang2015nonparametric}
Wang, X.; and Wang, Y. 2015.
\newblock Nonparametric multivariate density estimation using mixtures.
\newblock \emph{Statistics and Computing} 25(2): 349--364.

\bibitem[{Wang and Scott(2019)}]{wang2019nonparametric}
Wang, Z.; and Scott, D.~W. 2019.
\newblock Nonparametric density estimation for high-dimensional
  data—Algorithms and applications.
\newblock \emph{Wiley Interdisciplinary Reviews: Computational Statistics}
  11(4): e1461.

\bibitem[{Wei et~al.(2018)Wei, Lim, Zisserman, and Freeman}]{wei2018learning}
Wei, D.; Lim, J.~J.; Zisserman, A.; and Freeman, W.~T. 2018.
\newblock Learning and using the arrow of time.
\newblock In \emph{Proceedings of the IEEE Conference on Computer Vision and
  Pattern Recognition}, 8052--8060.

\bibitem[{Weston, Bengio, and Usunier(2010)}]{weston2010large}
Weston, J.; Bengio, S.; and Usunier, N. 2010.
\newblock Large scale image annotation: learning to rank with joint word-image
  embeddings.
\newblock \emph{Machine learning} 81(1): 21--35.

\bibitem[{Xiao et~al.(2015)Xiao, Xia, Yang, Huang, and Wang}]{xiao2015learning}
Xiao, T.; Xia, T.; Yang, Y.; Huang, C.; and Wang, X. 2015.
\newblock Learning from massive noisy labeled data for image classification.
\newblock In \emph{Proceedings of the IEEE conference on computer vision and
  pattern recognition}, 2691--2699.

\bibitem[{Xu et~al.(2017)Xu, Zhao, Xiao, Wu, Zhang, He, and
  Zhuang}]{xu2017video}
Xu, D.; Zhao, Z.; Xiao, J.; Wu, F.; Zhang, H.; He, X.; and Zhuang, Y. 2017.
\newblock Video question answering via gradually refined attention over
  appearance and motion.
\newblock In \emph{Proceedings of the 25th ACM international conference on
  Multimedia}, 1645--1653.

\bibitem[{Xu et~al.(2016)Xu, Mei, Yao, and Rui}]{xu2016msr}
Xu, J.; Mei, T.; Yao, T.; and Rui, Y. 2016.
\newblock Msr-vtt: A large video description dataset for bridging video and
  language.
\newblock In \emph{Proceedings of the IEEE conference on computer vision and
  pattern recognition}, 5288--5296.

\bibitem[{Yu, Kim, and Kim(2018)}]{yu2018joint}
Yu, Y.; Kim, J.; and Kim, G. 2018.
\newblock A joint sequence fusion model for video question answering and
  retrieval.
\newblock In \emph{Proceedings of the European Conference on Computer Vision
  (ECCV)}, 471--487.

\bibitem[{Zhang, Isola, and Efros(2016)}]{zhang2016colorful}
Zhang, R.; Isola, P.; and Efros, A.~A. 2016.
\newblock Colorful image colorization.
\newblock In \emph{European conference on computer vision}, 649--666. Springer.

\end{thebibliography}
\appendix
\onecolumn
\section*{Appendix Overview}
In Appendix \ref{proof_of_theorem} we present the proof of theorem \ref{theorem:error_bound}. In Appendix \ref{toy_dataset_appedix} we empirically reproduce the theoretical graphs in Section \ref{theoretical_analysis} using the multidimensional toy dataset described in Section \ref{toy_dataset} for corroborating the validity of the analysis presented in Thm. \ref{theorem:error_bound} in the multidimensional case. In Appendix \ref{qualitative_examples} we present qualitative examples. In Appendix \ref{low_t_failure_case_appendix} we present a visualization of the `low $T$ failure case' mentioned in Section \ref{numerical_simulations}. In Appendix \ref{unfair_evaluations} we compare the performance of our model trained without 3D features (i.e., only 2D) to the baseline model that is trained with 2D+3D features. In Appendix \ref{design_choice_analysis} we evaluate the effect of two important design choices. In Appendix \ref{additional_imp_details} we provide additional implementation details.
\section{Proof of Theorem \ref{theorem:error_bound}}
\label{proof_of_theorem}
\begin{proof}
    Given a pair $z_i=(x_i,y_i)$ with $(a_i,b_i)=(a,b)$, we find the moment generating function of $\Bar{S_i}$. 
    We first start with the moment generating functions of $s(x_{ik}, x_i)$ and $s(y_{ik}, y_i)$. We note that $-s(x_{ik}, x_i)$ and $-s(y_{ik}, y_i)$ are both distributed according to the folded normal distribution. For the general case, the moment generating function of $-|X|$, $X \sim \mathcal(\mu, \sigma)$, is
    \begin{equation}
        \label{eq:mgf_minus_abs_A}
        \mathcal{M}(t; \mu, \sigma) = e^{\frac{\sigma^2 t^2}{2} + \mu t}\Phi\bigg(\frac{\mu}{\sigma} + \sigma t\bigg) + 
        e^{\frac{\sigma^2 t^2}{2} - \mu t}\Phi\bigg(-\frac{\mu}{\sigma} + \sigma t\bigg),
    \end{equation}
    where $\Phi(\cdot)$ is the normal cumulative distribution function.
    
    Denote the cardinality of the set of pairs that originate from the components $(a,b)$ of $z_i$ as $\tilde{m}_i$, i.e., $\tilde{m}_i \triangleq \vert\{z_j: a_j=a, b_j=b, j=1,\dots,M\}\vert$. Recall the three sigma limit assumption between each pair of components. Thus, by using the law of total expectation, the moment generating function of $\Bar{S}_i \given p_i$ can be expressed as
    \begin{align}
        \label{eq:mgf_S_i_bar}
        \mathcal{M}_{\Bar{S}_i \given p_i}(t) & = \sum_{n=0}^{M}{Pr(\tilde{m}_i = n \given p_i)\cdot \mathcal{M}_{\Bar{S}_i \given p_i, \tilde{m}_i = n}(t)},
    \end{align} 
    where, 
    \begin{multline}
            \mathcal{M}_{\Bar{S}_i \given \tilde{m}_i = n}(t) = \bigg[\mathcal{M}\bigg(\frac{1}{2K}t; \mu_{a} - x_i, \sigma_{a}\bigg)\cdot
            \mathcal{M}\bigg(\frac{1}{2K}t; \mu'_{b} - y_i, \sigma'_{b}\bigg)\bigg]^{\min\{n, K\}}
            \cdot \\
            \prod_{j=1}^{\max\{0, K-n\}}{\mathcal{M}\bigg(\frac{1}{2K}t; \mu_{\alpha_j} - x_i, \sigma_{\alpha_j}\bigg)}
            \cdot \mathcal{M}\bigg(\frac{1}{2K}t; \mu'_{\beta_j} - y_i, \sigma'_{\beta_j}\bigg),
    \end{multline}    
    where all $(\alpha_j, \beta_j) \ne (a,b)$, and
    \begin{equation}
        Pr(\tilde{m}_i = n \given p_i) = 
        \begin{cases}
            \binom{M}{n} \cdot \big(\frac{1-\eta}{T}\big)^n \cdot \big(1 - \frac{1-\eta}{T}\big)^{M-n} & p_i = 1\\
            \binom{M}{n} \cdot \big(\frac{\eta}{T(T-1)}\big)^n \cdot \big(1 - \frac{\eta}{T(T-1)}\big)^{M-n} & p_i = 0.
        \end{cases}
        \label{eq:pr_m_i_tilde}
    \end{equation}
    We used the fact that the moment generating function of $V = c_1 U_1 + \cdots + c_n U_n$ is given by $\mathcal{M}_V(t) = \mathcal{M}_{U_1}(c_1 t) \cdots \mathcal{M}_{U_n}(c_n t)$, when the $c_i$'s are scalars and the $U_i$'s are independent random variables.
    
    Applying the Chernoff bound concludes the proof.
\end{proof}

To make sense of this error bound we performed multiple numerical simulations (See Section \ref{numerical_simulations}), with the following set up:
\begin{enumerate}
    \item $\{\mu_i, \mu'_i\}_{i=1}^T$ are sampled uniformly from $[0, 100]$
    \item $\{\sigma_i\}_{i=1}^T$ are sampled uniformly from $(0, \displaystyle{\min_{i, j}{|\mu_i - \mu_j|}} / 6]$, and $\{\sigma'_i\}_{i=1}^T$ are sampled uniformly from $(0, \displaystyle{\min_{i, j}{|\mu'_i - \mu'_j|}} / 6]$, such that the three sigma limit assumption in Theorem \ref{theorem:error_bound} is met.
    \item We set $\tau=\tau^*$ such that $P(\Bar{S_i} \geq \tau^* \given p_i = 0) = P(\Bar{S_i} \leq \tau^* \given p_i = 1)$, i.e., $\tau^* = \sqrt{\mathcal{M}_{\Bar{S}_i \given p_i = 0}(t) \cdot \mathcal{M}_{\Bar{S}_i \given p_i = 1}(-t)}$.
    \item Optimization for $t$ is done over $[1, 2, ..., 100]$.
    \item Each experiment is repeated 10 times and the average is presented. 
\end{enumerate}
\clearpage
\section{Reproducing Empirically the Theoretical Results}
\label{toy_dataset_appedix}
\begin{figure*}[ht]
    \centering
    \begin{subfigure}[t]{0.33\textwidth}
        \centering
        \includegraphics[width=1\linewidth]{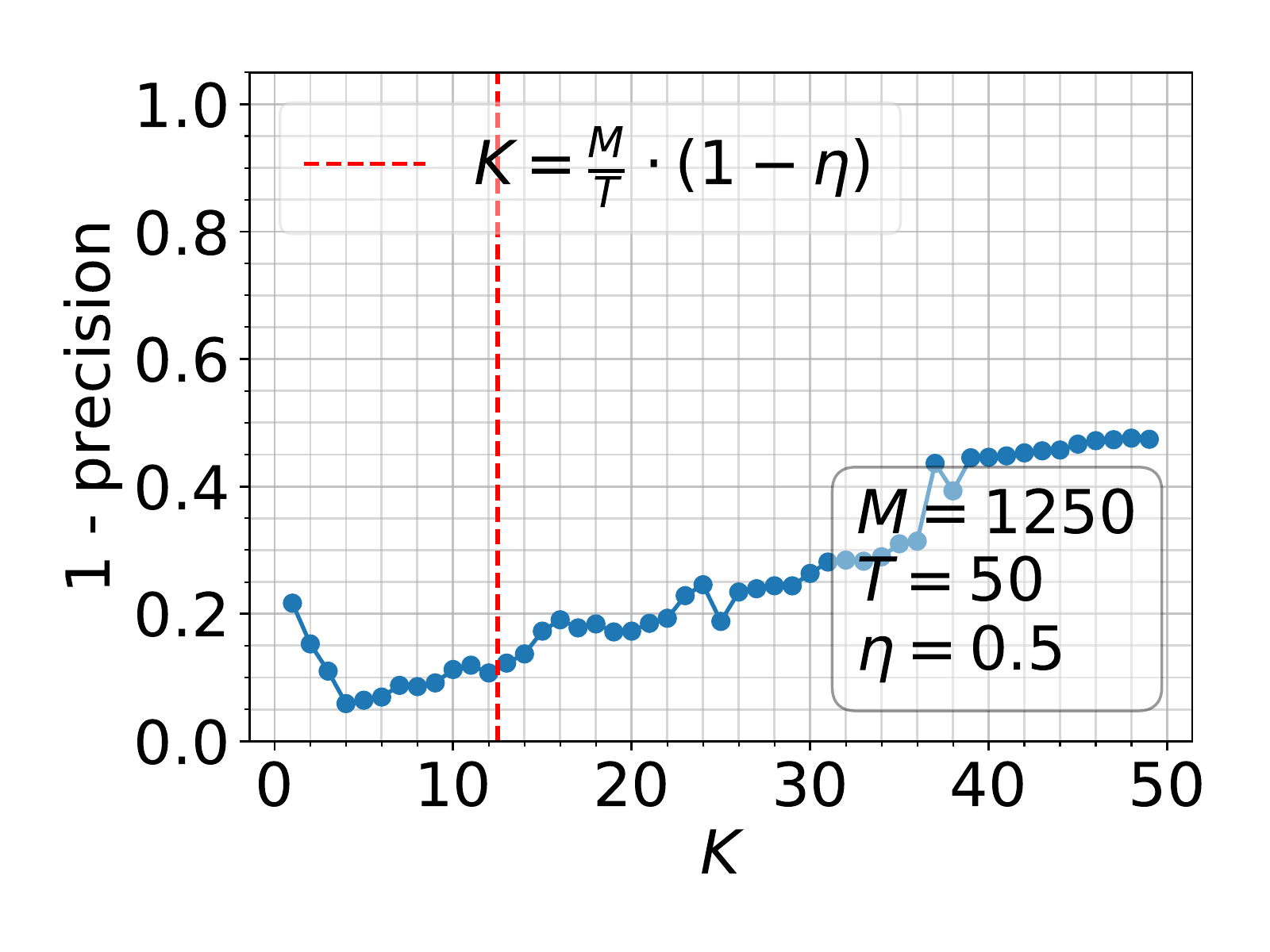}
        \caption{$K$}
        \label{fig:toy_K_sweep}
    \end{subfigure}\hfill
    \begin{subfigure}[t]{0.33\textwidth}
        \centering    
        \includegraphics[width=1\linewidth]{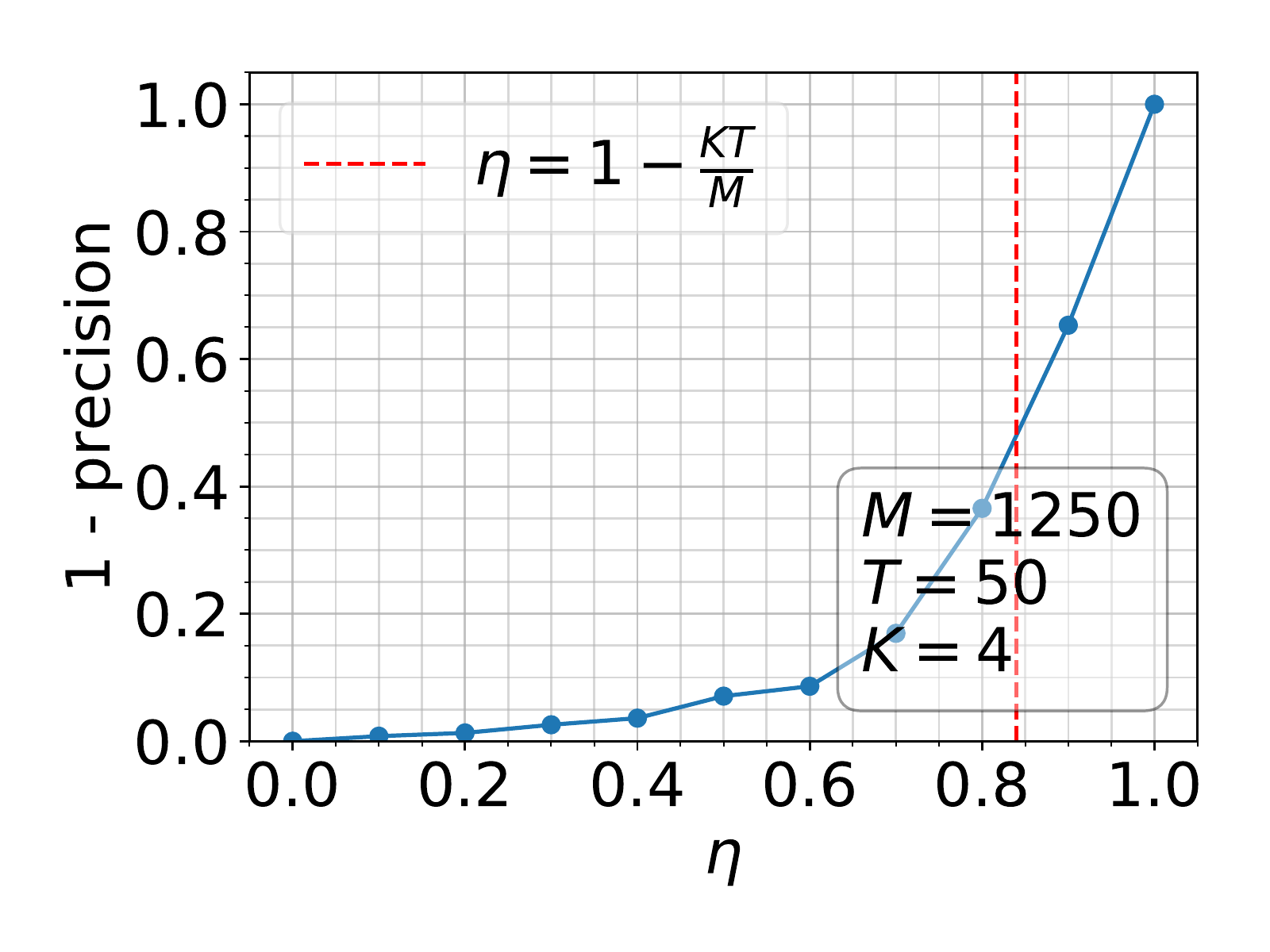}
        \caption{$\eta$}
        \label{fig:toy_eta_sweep} 
    \end{subfigure}\hfill
    \begin{subfigure}[t]{0.33\textwidth}
        \centering    
        \includegraphics[width=1\linewidth]{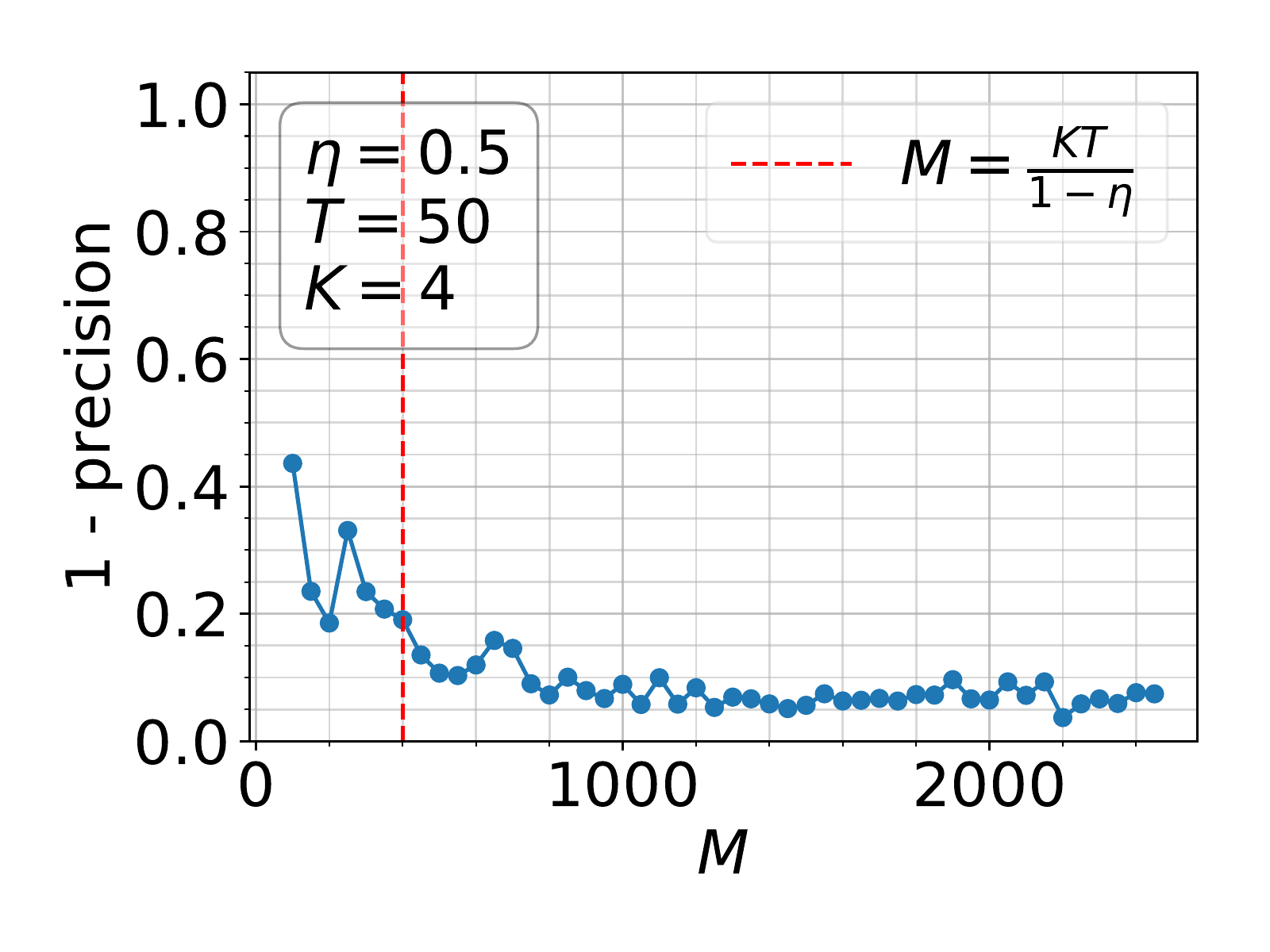}
        \caption{$M$}
        \label{fig:toy_M_sweep} 
    \end{subfigure}\hfill
    \begin{subfigure}[t]{0.33\textwidth}
        \centering
        \includegraphics[width=1\linewidth]{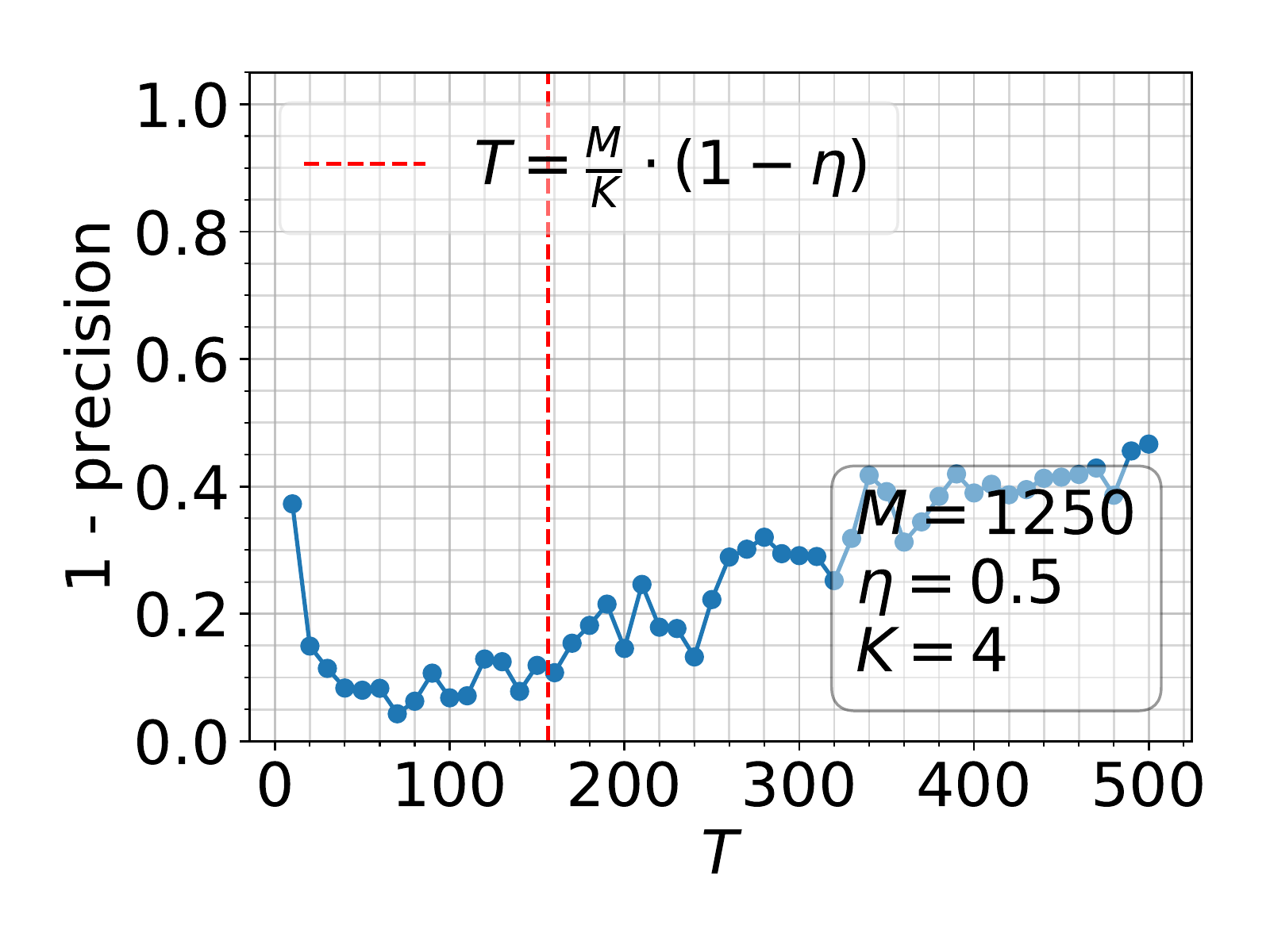}
        \caption{$T$}
        \label{fig:toy_T_sweep}
    \end{subfigure}\hfill
    \begin{subfigure}[t]{0.33\textwidth}
        \centering
        \includegraphics[width=1\linewidth]{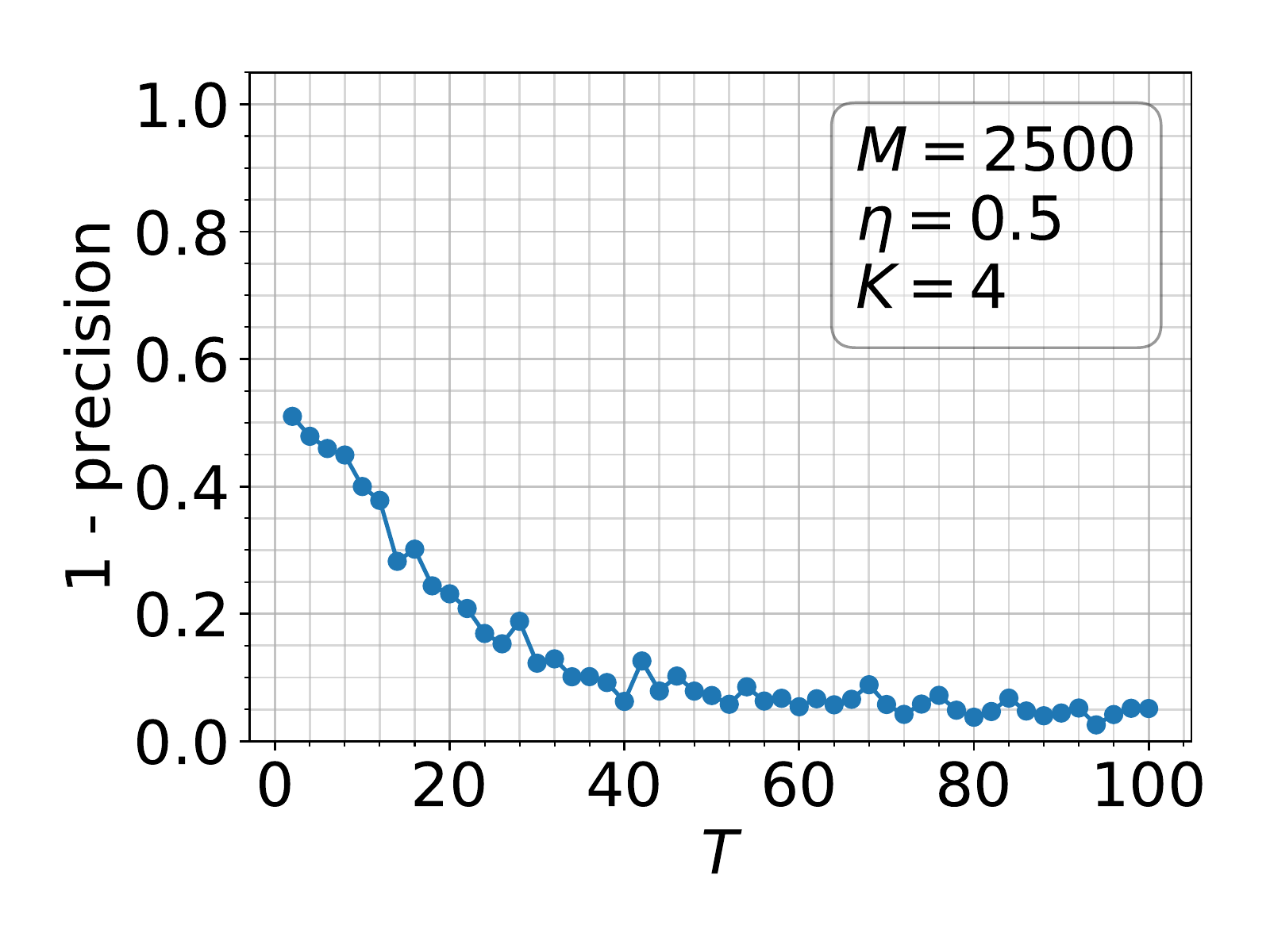}
        \caption{Low $T$ failure case}
        \label{fig:toy_low_t_failure_case}
    \end{subfigure}
    \caption{\textbf{Empirical error of multidimensional toy dataset}. In the figures above we empirically reproduce the theoretical graphs in Fig. \ref{fig:numerical_simulations} and Fig. \ref{fig:low_t_failure_case} using the multidimensional toy dataset mentioned above. The results help in validating that the analysis that was done in Thm. \ref{theorem:error_bound} for a single dimension, also holds for a multidimensional space. The vertical axis is $(1 - \mathrm{precision})$, where the threshold was chosen such that F1 Score is maximized. In each figure we mark with a red dashed vertical line the point at which the equation $K = \frac{M}{T} \cdot (1-\eta)$ holds. We note that across all figures, similar trends are observed in comparison to the theoretical graphs in Fig. \ref{fig:numerical_simulations} and Fig. \ref{fig:low_t_failure_case}.}
    \label{fig:empirical_exp}
\end{figure*}

\begin{figure}[ht]
    \centering    
    \includegraphics[width=0.5\linewidth]{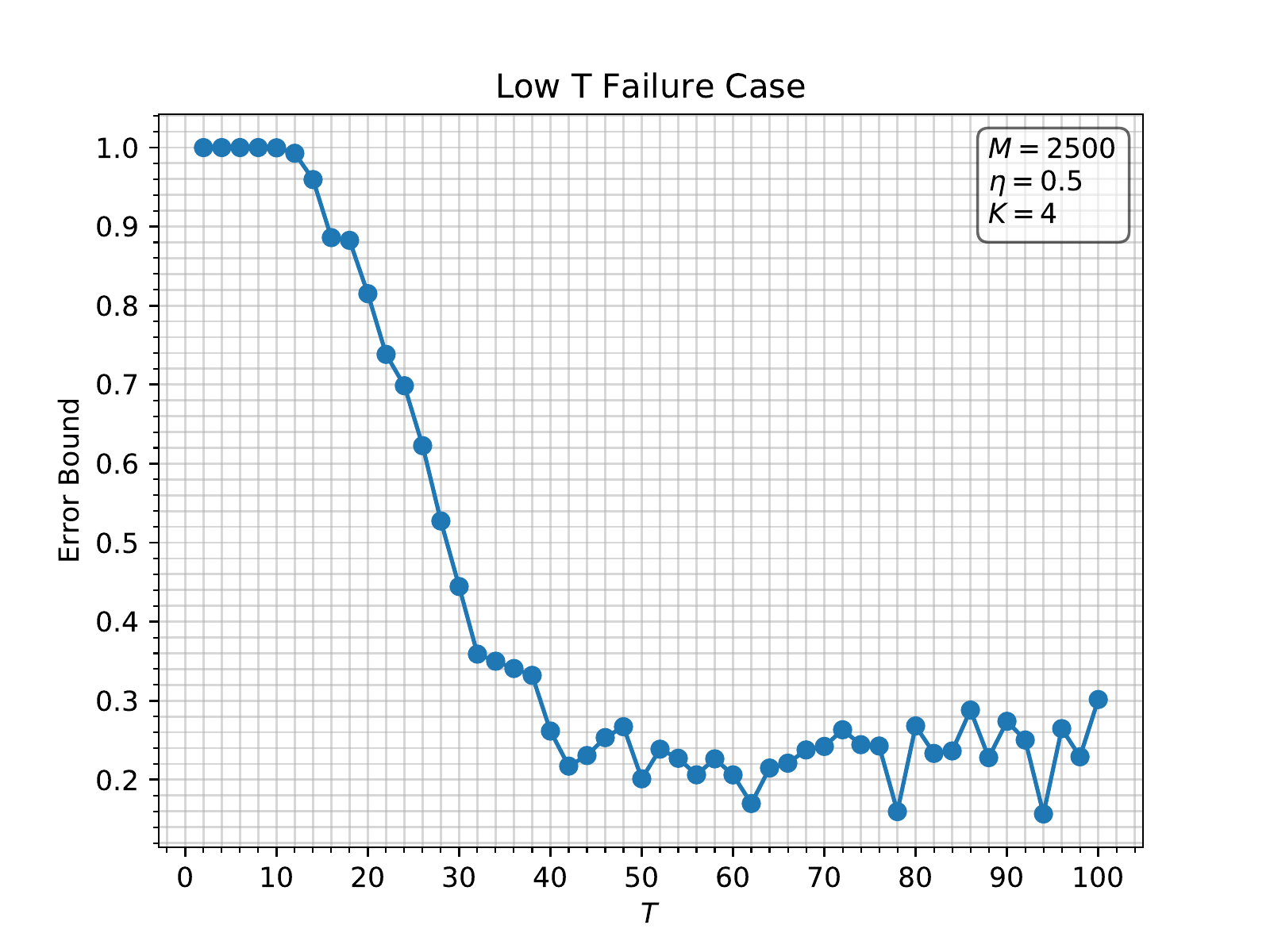}
    \caption{Low $T$ failure case.}
    \label{fig:low_t_failure_case} 
\end{figure}

In Fig. \ref{fig:empirical_exp}  we empirically reproduce the theoretical graphs in Section \ref{theoretical_analysis} Fig. \ref{fig:numerical_simulations} and Fig. \ref{fig:low_t_failure_case} using the multidimensional toy dataset described in Section \ref{toy_dataset}. These results corroborate the validity of the analysis presented in Thm. \ref{theorem:error_bound} in the multidimensional case. We note that across all sub-figures in Fig. \ref{fig:empirical_exp}, similar trends are observed in comparison to the theoretical graphs in Fig. \ref{fig:numerical_simulations} and Fig. \ref{fig:low_t_failure_case}. More specifically, across Fig. \ref{fig:toy_K_sweep}, \ref{fig:toy_eta_sweep}, \ref{fig:toy_M_sweep}, \ref{fig:toy_T_sweep} we observe the influence of $K_0$ (mentioned in Section \ref{theoretical_analysis}) on the performance of our suggested approach. Additionally, in Fig. \ref{fig:toy_low_t_failure_case}, we observe the same `low T failure case' mentioned in Section \ref{theoretical_analysis}.
\clearpage
\section{Qualitative Examples}
\label{qualitative_examples}
\subsection{High Score Examples and Their {$\boldsymbol{k}$} Nearest Neighbours}
\begin{figure*}[ht]
    \centering    
    \includegraphics[width=0.47\linewidth]{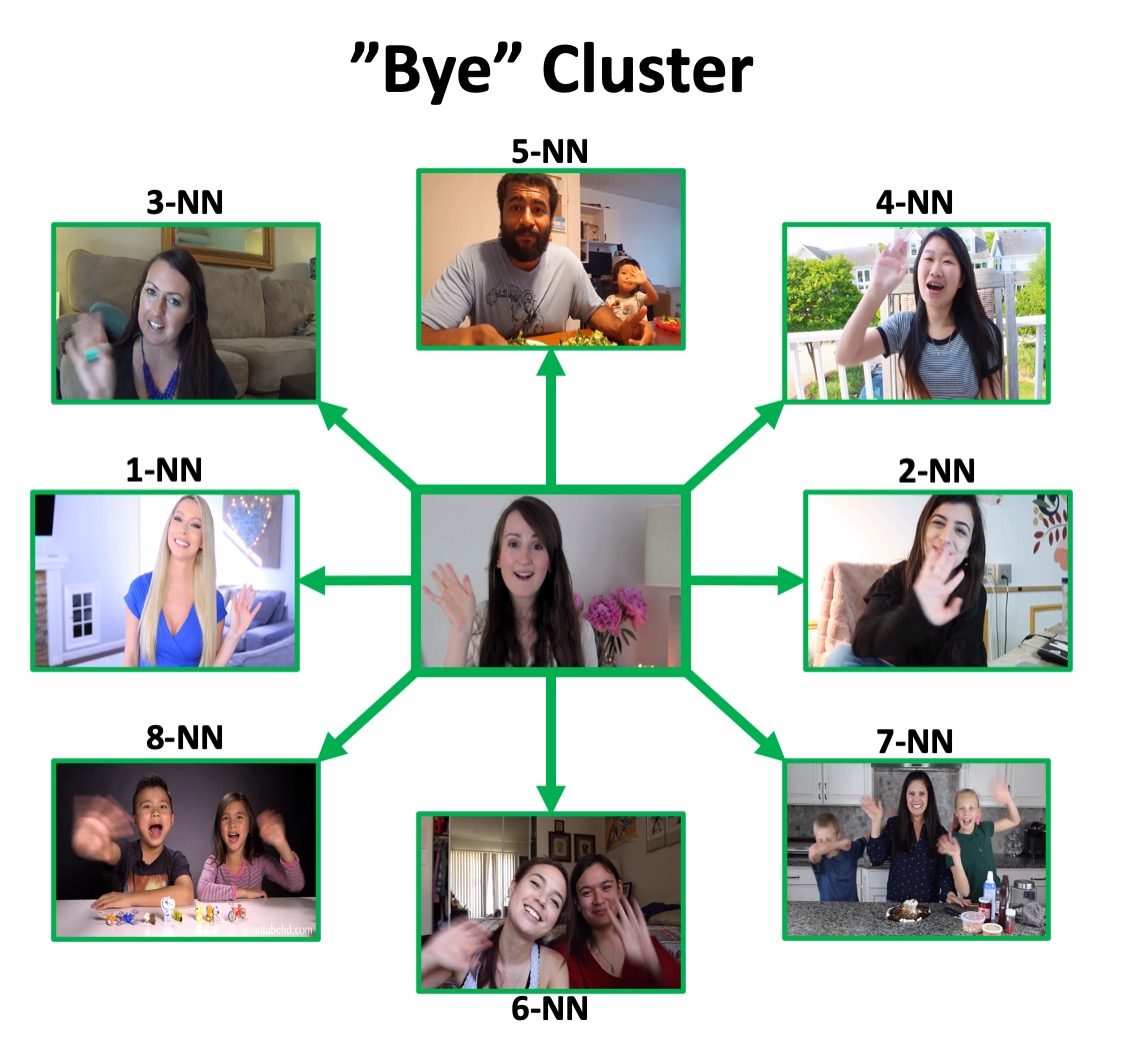}
    \caption{`Bye' Cluster.}
    \label{fig:bye_cluster} 
\end{figure*}

\begin{figure*}[ht]
    \centering    
    \includegraphics[width=0.47\linewidth]{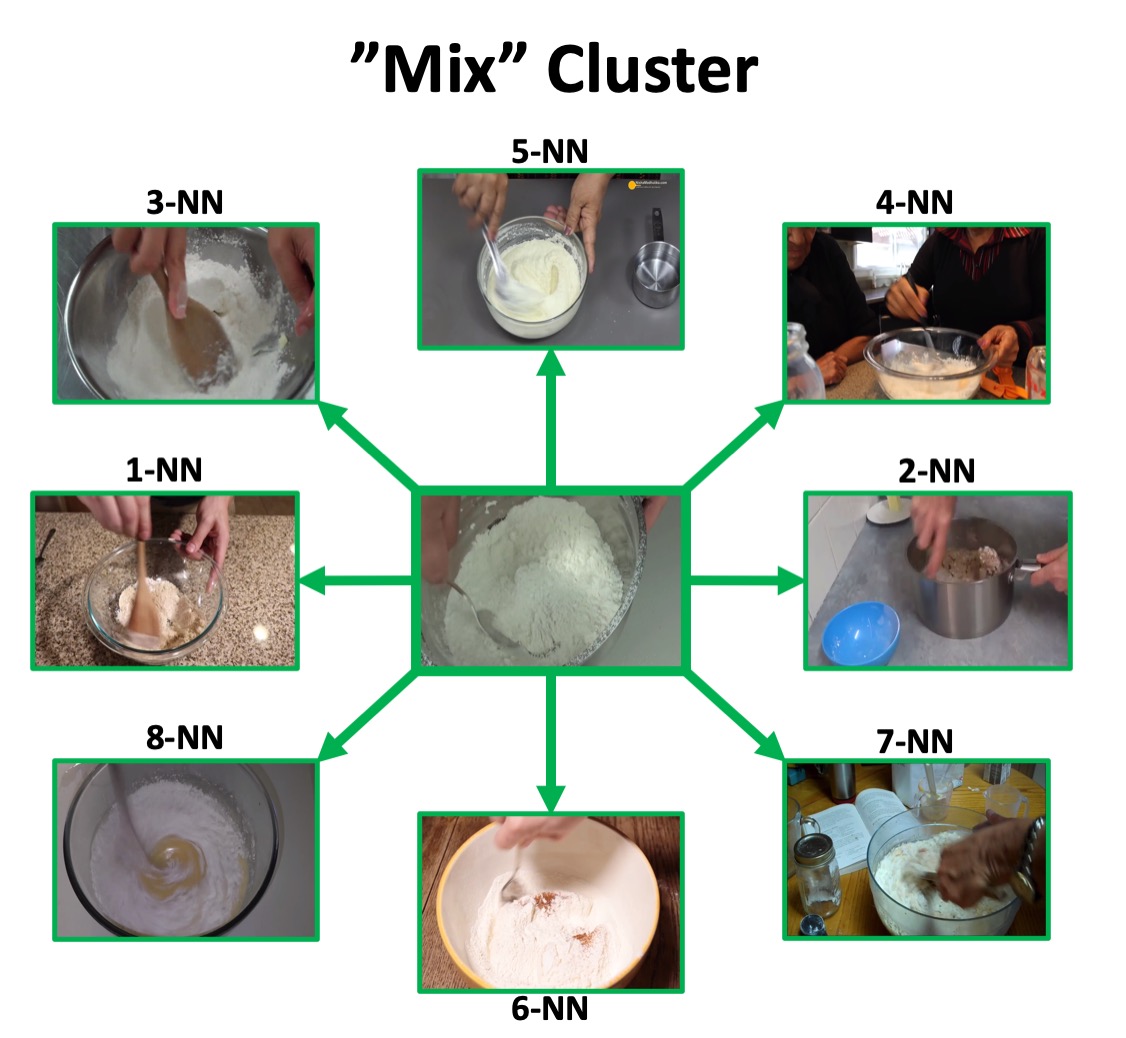}
    \caption{`Mix' Cluster}
    \label{fig:mix_cluster} 
\end{figure*}

In this section, we present two examples of clip-caption pairs with high $\hat{p}$ (see Eq. \eqref{eq:p_i}) and their nearest neighbours in the multimodal space that contributed to their high score. One example includes in its caption the word `bye' (Fig. \ref{fig:bye_cluster}) and the other the word `mix' (Fig \ref{fig:mix_cluster}). The figures below include only one representative frame from each video clip. Thus, it is worth mentioning that all video clips in Fig. \ref{fig:bye_cluster} and \ref{fig:mix_cluster} in fact contain a `waiving goodbye' action and a `mixing' action, respectively. It is important to note that these examples were extracted only based on our noise estimation component and are not based on the learned shared embedding space.
\clearpage
\subsection{High and Low Score Examples for the Same Query}
In this section, we present ten examples of high and low $\hat{p}$ score clip-caption pairs for the same query. These examples visually illustrate how our noise estimation component is able to distinguish between {\it wrongly associated} pairs and {\it correctly associated} pairs successfully.
\begin{figure}[ht]
    \centering    
    \includegraphics[width=0.60\linewidth]{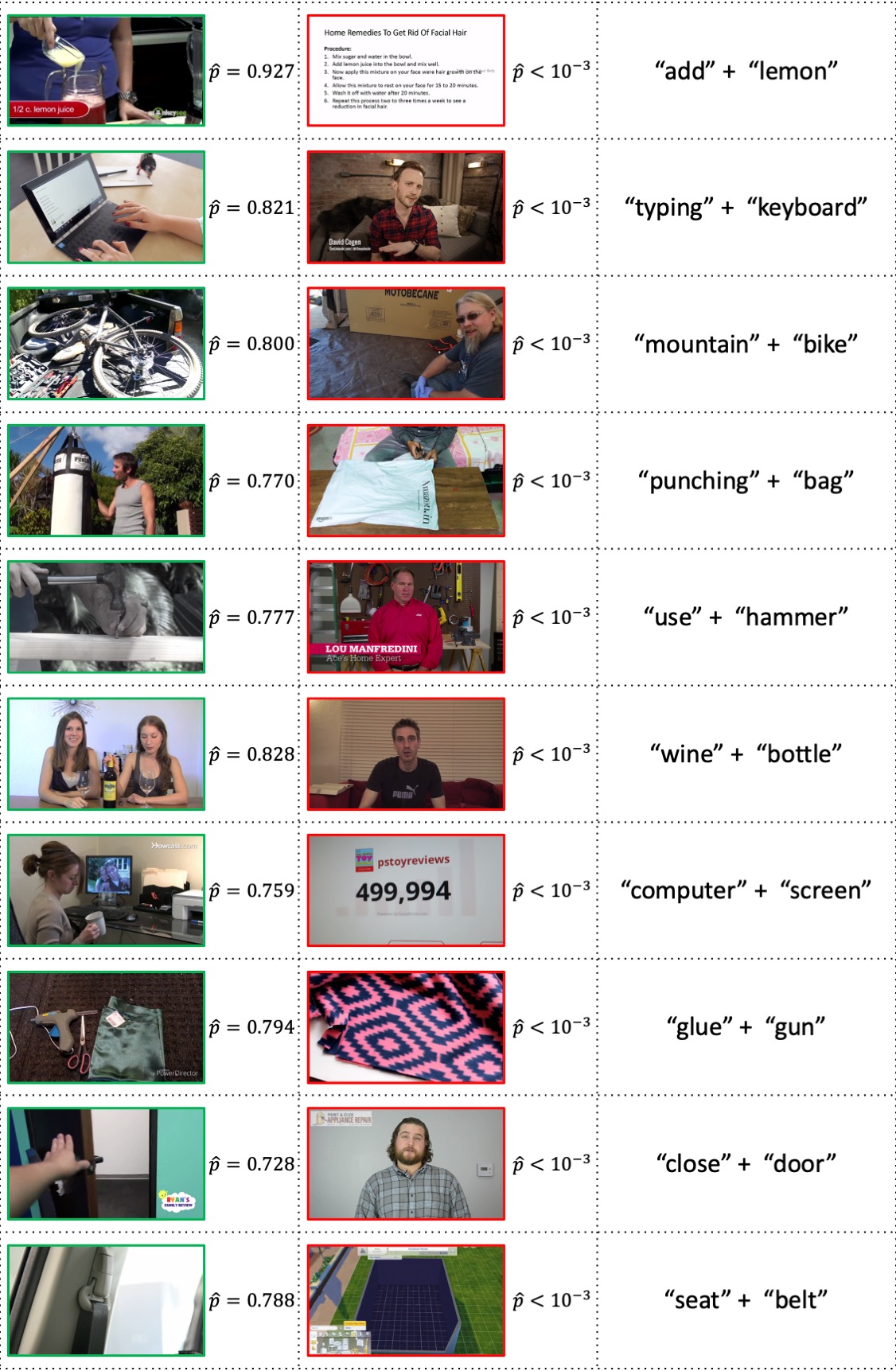}
    \caption{High and low $\hat{p}$ score examples for the same query. Each row contains two video clips (represented by a single frame) that include in their caption the same query (right column). The left column (GREEN) contains clips with high $\hat{p}$, while the middle column (RED) contains clips with low $\hat{p}$.}
    \label{fig:high_low_pairs} 
\end{figure}
\clearpage
\section{Low {$\boldsymbol{T}$} Failure Case Simulation}
\label{low_t_failure_case_appendix}
One instance where the method fails, gives us insight into why usually the method succeeds. The method will fail for a small number of concepts ($T$) and regardless of $K_0$, because for a small number of concepts there is a higher chance that two or more {\it wrongly associated} pairs belong to the same pair of concepts in both modalities. Fortunately, in real-world data $T$ is almost always large, and additionally as $T$ increases this problem is alleviated by a factor of $\mathcal{O}(T^2)$ (see Appendix \ref{proof_of_theorem}, Eq. \eqref{eq:pr_m_i_tilde}). A simulation of this case is presented in the figure below.
\label{low_t_failure_case}

\section{Zero-shot Text-To-Video retrieval in `unfair' settings}
\label{unfair_evaluations}
We demonstrate that our model outperforms or is at least on par with the performance of HTM-PT \cite{miech2019howto100m} even given a setting which is a clear disadvantage such as training it without 3D features (i.e., only 2D). See Table \ref{table:unfair_evaluations} in Appendix \ref{unfair_evaluations}. This shows:
    \begin{enumerate}[label=(\roman*)]
        \item The power of our noise estimation method and its potential.
        \item Integrating our multimodal density estimation component allows saving time and/or computation power by training and running inference with only 2D features, without (or with minor) performance degradation.
    \end{enumerate}
We note that specifically for MSRVTT dataset our 2D-based model actually performs slightly better than our 2D+3D-based model. This result requires further investigation.
\begin{table*}[ht]
\small
  \centering
  \caption{Zero-shot Text-To-Video retrieval in `unfair' settings. For MR the lower the better. We show below that our model outperforms or is at least on par with the performance of HTM-PT \cite{miech2019howto100m} even given a setting which is a clear disadvantage such as training it without 3D features (no-3D), i.e., only 2D features}
  \setlength\tabcolsep{3pt}
  \begin{tabular}{lccccrccccrcccc}
    \toprule
          & \multicolumn{4}{c}{MSRVTT} && \multicolumn{4}{c}{LSMDC} && \multicolumn{4}{c}{MSVD} \\ 
    \cmidrule{2-5} \cmidrule{7-10} \cmidrule{12-15}
    Method & R@1 & R@5 & R@10 & MR && R@1 & R@5 & R@10 & MR && R@1 & R@5 & R@10 & MR \\ 
    \midrule
    HTM-PT$^*$ \cite{miech2019howto100m} & 7.5 & 21.2 & 29.6 & 38.0 && \textbf{4.0} & 9.8 & 14.0 & 137.0 && \textbf{12.86} & 33.06 & \textbf{45.83} & \textbf{13.0} \\
    HTM-PT \cite{miech2019howto100m} no-3D & 6.9 & 19.8 & 27.4 & 43.0 && 3.3 & 9.9 & 13.4 & 147.0 && 11.57 & 30.25 & 40.84 & 17.0\\
    Ours (no 3D) & \textbf{8.4} & \textbf{22.0} & \textbf{30.4} & \textbf{36.0} && \textbf{4.0} & \textbf{10.5} & \textbf{14.3} & \textbf{141.5} && 12.74 & \textbf{33.48} & 44.96 & 14.0 \\
    \bottomrule
  \end{tabular}
  \label{table:unfair_evaluations}
\end{table*}
\clearpage
\section{Design-Choice Analysis}
\label{design_choice_analysis}
In this section, we evaluate the effect of two important design choices on the Zero-Shot Text-To-Video Retrieval task: (a) $k$-NN parameter; and (b) $S(\cdot, \cdot)$, the multimodal similarity function. In Table \ref{table:design_choice_analysis_k_param} ($K$ analysis) we see a similar trend as in Section \ref{theoretical_analysis}, Fig. \ref{fig:numerical_simulations}a, i.e., increasing $K$ decreases the error initially and from a certain value, the error increases. In Table \ref{table:design_choice_analysis_sim_functions} ($S(\cdot, \cdot)$ analysis), it is evident that there is a slight advantage of using the minimum function over the mean function, yet it is not conclusive, i.e., the mean function that is used in Thm. \ref{theorem:error_bound} for simplicity of analysis is a decent design choice as well.
\begin{table}[ht]
\small
    \begin{subtable}[h]{0.49\textwidth}
      \centering
      \setlength\tabcolsep{3pt}
      \begin{tabular}{lccc}
        \toprule
        $k$-NN & MSRVTT & LSMDC & MSVD\\ 
        \midrule
        1-NN & 20.9 & 10.7 & 33.66 \\
        4-NN & \textbf{21.3} & \textbf{11.6} & \textbf{35.7} \\
        16-NN & 20.8 & 11.3 & 35.55 \\
        \bottomrule
      \end{tabular}
      \caption{$k$-NN}
      \label{table:design_choice_analysis_k_param}
    \end{subtable}
    \begin{subtable}[h]{0.49\textwidth}
        \centering
      \setlength\tabcolsep{3pt}
      \begin{tabular}{lccc}
        \toprule
        $S(z_i, z_j)$ & MSRVTT & LSMDC & MSVD\\ 
        \midrule
        mean (used in Thm. \ref{theorem:error_bound}) & 20.3 & 11.2 & \textbf{35.84} \\
        minimum (Eq. \eqref{eq:dist}) & \textbf{21.3} & \textbf{11.6} & 35.7\\
        \bottomrule
      \end{tabular}
      \caption{Multimodal similarity function}
      \label{table:design_choice_analysis_sim_functions}
     \end{subtable}
\label{table:design_choice_analysis}
\caption{Design-Choice Analysis. Recall@5 results for Zero-Shot Text-To-Video Retrieval.}
\end{table}

\section{Additional Implementation Details}
\label{additional_imp_details}
\textbf{Sampling strategy.} For a fair comparison to the baseline model HTM \cite{miech2019howto100m}, we follow the same sampling strategy. More specifically, half of the negative pairs, $(v_i, c_j): i \neq j$ are sampled such that they belong to the same video clip, while the other half are sampled such that they do not.
\\\\
\textbf{$k$-NN Computation.} To compute $k$-NN efficiently over the entire dataset we use FAISS \cite{johnson2019billion}. Due to the high correlation between video segments of the same video, in practice we extract $K$ nearest neighbors that originate from different videos, where $K = 4$. 

\begin{table}[ht]
\small
    \begin{subtable}[h]{0.5\textwidth}
      \centering
      \setlength\tabcolsep{2pt}
      \begin{tabular}{lccc}
        \toprule
        Dataset & Length [s] & \#Clips & Train/Val/Test Split\\ 
        \midrule
        MSRVTT-QA \shortcite{xu2017video} & 20 & 10,000 & 158,581/12,278/72,821\\
        MSVD-QA \shortcite{xu2017video} & 20 & 1970 & 30,933/6,415/13,157\\
        \bottomrule
      \end{tabular}
      \caption{Video question answering}
      \label{table:vqa_stats}
    \end{subtable}
    \hfill
    \begin{subtable}[h]{0.5\textwidth}
      \centering
      \setlength\tabcolsep{2pt}
      \begin{tabular}{lccc}
        \toprule
        Dataset & Length [s] & \#Clips & Train/Val/Test Split\\ 
        \midrule
        MSRVTT \shortcite{xu2016msr} & 20 & 10,000 & 130,260/9940/1000 \\
        MSVD \shortcite{chen2011collecting} & 20 & 1970 & 48,820/4401/3350\\
        LSMDC \shortcite{rohrbach2015dataset} & 5 & 128,085 & 101,079/7408/1000\\
        \bottomrule
      \end{tabular}
      \caption{Text-Video retrieval}
      \label{table:retrieval_stats}
     \end{subtable}
\caption{Statistics of datasets. For retrieval we use the same test set split as defined by \cite{mithun2018learning,yu2018joint,miech2018learning} for a fair comparison.}
\end{table}

\end{document}